\documentclass[10pt,twocolumn,letterpaper]{article}
\usepackage{iccv}      % To produce the REVIEW version
\usepackage[accsupp]{axessibility}
\usepackage{xcolor}
\usepackage{float}
\usepackage{breqn}
\usepackage{flexisym}
\usepackage{mathtools}
\usepackage{multirow}
\usepackage{array}
\usepackage{makecell}
\usepackage{natbib}
\usepackage{stfloats}
\usepackage{algorithm}
\usepackage{algorithmic}
\usepackage{newfloat}
\usepackage{listings}
\usepackage{amsmath}
\usepackage{diagbox}
\usepackage{colortbl}
\usepackage{trimclip}
\definecolor{dgreen}{rgb}{0.13,0.7,0.30}
\definecolor{purple}{rgb}{0.64,0.29,0.64}

\newcommand{\black}{\color{black}}
\newcommand{\vnsp}{\vspace{-0.1in}}
\definecolor{iccvblue}{rgb}{0.21,0.49,0.74}
\usepackage[pagebackref,breaklinks,colorlinks,allcolors=iccvblue]{hyperref}

\def\paperID{8503} % *** Enter the Paper ID here
\def\confName{ICCV}
\def\confYear{2025}

\title{Intervening in Black Box: Concept Bottleneck Model for Enhancing Human Neural Network Mutual Understanding}

\author{Nuoye Xiong\textsuperscript{1}, Anqi Dong\textsuperscript{2}\footnotemark[1], Ning Wang\textsuperscript{1}, Cong Hua\textsuperscript{1}, Guangming Zhu\textsuperscript{1},\\
	Lin Mei\textsuperscript{3}, Peiyi Shen\textsuperscript{1}, Liang Zhang\textsuperscript{1}\footnotemark[2]\\
	\textsuperscript{1}Xidian University, China \\
	\textsuperscript{2}KTH Royal Institute of Technology, Sweden \\
	\textsuperscript{3}Donghai Laboratory, China \\
	{\tt\small \{nyx, ningwang, chua\}@stu.xidian.edu.cn} \\
	{\tt\small \{gmzhu, pyshen, liangzhang\}@xidian.edu.cn} \\
	{\tt\small \{meilin\}@donghailab.com} \\
	{\tt\small \{anqid\}@kth.se}
}

\begin{document}
	\setlength{\belowdisplayskip}{8pt}
	\setlength{\abovedisplayskip}{8pt}
	\setlength{\belowdisplayshortskip}{0pt}
	\setlength{\abovedisplayshortskip}{0pt}
	
	\maketitle

	\renewcommand{\thefootnote}{\fnsymbol{footnote}} 
	\footnotetext[1]{Co-first contributing author.} 
	\footnotetext[2]{Corresponding author.} 
	
	\begin{abstract}
		Recent advances in deep learning have led to increasingly complex models with deeper layers and more parameters, reducing interpretability and making their decisions harder to understand. While many methods explain black-box reasoning, most lack effective interventions or only operate at sample-level without modifying the model itself. To address this, we propose the Concept Bottleneck Model for Enhancing Human-Neural Network Mutual Understanding (CBM-HNMU). CBM-HNMU leverages the Concept Bottleneck Model (CBM) as an interpretable framework to approximate black-box reasoning and communicate conceptual understanding. Detrimental concepts are automatically identified and refined (removed/replaced) based on global gradient contributions. The modified CBM then distills corrected knowledge back into the black-box model, enhancing both interpretability and accuracy. We evaluate CBM-HNMU on various CNN and transformer-based models across Flower-102, CIFAR-10, CIFAR-100, FGVC-Aircraft, and CUB-200, achieving a maximum accuracy improvement of $2.64\%$ and a maximum increase in average accuracy across $1.03\%$. Source code is available at: {\color{blue} https://github.com/XiGuaBo/CBM-HNMU}.
	\end{abstract}
	\black

	\section{Introduction}\label{sec:intro}
	Interpretable machine learning has become a crucial research direction, particularly in image classification, where understanding model decisions is essential for trust and reliability. Neural networks for image classification can be broadly categorized into three types by their interpretability: (i) Black-box models, where decision-making is opaque and difficult to trace; (ii) Gray-box models, which offer partial transparency, allowing some explanation of decisions; and (iii) White-box models, which are fully transparent with completely traceable decision processes \cite{Guidotti2018ASO, Ali2023ExplainableAI}. While black-box models often achieve superior performance, their lack of interpretability poses significant challenges.
	
	Many studies aim to achieve white-box-like interpretability while maintaining black-box models' performance. Methods such as Grad-CAM \cite{Selvaraju2016GradCAMVE}, Saliency Maps \cite{Simonyan2013DeepIC}, and Feature Attack \cite{Engstrom2019AdversarialRA} use \emph{feature attribution}, helping to explain its focus and reasoning in making predictions. Alternatively, methods include ACE \cite{Ghorbani2019TowardsAC}, MOCE \cite{Kim2024WhatDA}, IBD \cite{Zhou2018InterpretableBD}, and CRAFT \cite{Fel2022CRAFTCR} focus on \emph{visual concept} explanations, linking model decisions to human-understandable patterns or attributes in data. These approaches provide more structured interpretation by associating model reasoning with recognizable concepts instead of abstract features. However, most existing methods focus on passive explanations rather than allowing active intervention to modify black-box behavior. This limitation weakens their ability to systematically refine model decision-making.

	The Concept Bottleneck Model (CBM) \cite{Koh2020ConceptBM} introduced an intervenable framework for concept-based reasoning, expanding the scope of interpretability methods. However, its reliance on \textit{concept bottleneck} representations often leads to lower classification accuracy, compared to black-box models with the same backbone. 
	Post-hoc CBM \cite{Yuksekgonul2022PosthocCB} was later introduced to mitigate this limitation by integrating black-box residual connections. Yet, this modification weakens overall interpretability and, even with interventions, Post-hoc CBM still falls short of achieving the classification accuracy of original black-box model. More critically, its intervention mechanism is in the same vein as CBM and is constrained to adjusting the concept weight matrix, that heavily dependent on human priors. As a result, biases inherent in the backbone \cite{zhou2025mitigating} cannot be effectively corrected.

	Beyond CBM-based approaches, numerous studies employ interpretable surrogate structures and human-understandable concept space to explain black-box behavior without compromising performance \cite{Ge2021API, Moreira2024DiConStructCC}. While these methods enable transparent reasoning, answering questions like ``{\it Why is it/is it not}'', they focus on explanation rather than intervention. Beyond CBM \cite{Marcinkevics2024BeyondCB} addresses this limitation by using probe functions to map intermediate black-box representations to a human-understandable concept space. This approach identifies biases and aligns reasoning pathways by modifying the concept distribution based on prior knowledge. However, its interventions are restricted to the sample level, without directly modifying black-box parameters, and require additional data to train the probe function for concept space projection.

	Despite advances in interpretability of black-box models, existing methods primarily diagnose errors but offer limited mechanisms for correction. To address this gap, we propose the Concept Bottleneck Model for Enhancing Human Neural Network Mutual Understanding (CBM-HNMU). The contributions can be summarized as
	\begin{enumerate}
		\item {\bf Integrated Explanation Framework:} CBM-HNMU integrates feature attribution and multimodal concept-based explanations to enhance interpretability in black-box inference. It aligns CBM with black-box hidden layer concepts and class predictions for more transparent reasoning and less approximation bias.
		\item {\bf Beyond Sample-Level Automatic Intervention:} CBM-HNMU introduces black-box model modification and error correction beyond individual samples. By computing global concept contributions based on class-level gradient, it autonomously identifies detrimental concepts for CBM intervention. Knowledge in intervention refines black-box model's parameters through knowledge distillation, correcting classification errors.
		
		\item {\bf Label-free Human-AI Mutual Understanding:} CBM-HNMU utilizes natural language concept bottlenecks generated by ChatGPT-3.5-Turbo \cite{Brown2020LanguageMA} and cross-modal probe based on OpenAI-CLIP \cite{Radford2021LearningTV} to enhance interpretability and performance without requiring additional human annotations.
	\end{enumerate}

	\section{Related Works}\label{sec:rela}
	
	\subsection{Model Approximation}
	Model approximation employs external interpretable models to replicate and explain black-box decision-making, offering insights without directly modifying original model. For example, \citet{Ge2021API} utilized graph neural networks and visual concepts from ACE to develop the Visual Reasoning Explanation Framework (VRX). This framework facilitates human-to-network interaction through Knowledge Distillation (KD) \cite{Hinton2015DistillingTK}, addressing the question, ``Why is it?'' in model decisions. Similarly, DiConStruct \cite{Moreira2024DiConStructCC} employs a distillation-based framework that integrates an approximate black-box graph reasoning network as a surrogate model. This approach efficiently approximates black-box predictions while providing causal explanations \cite{Beckers2022CausalEA, Warren2022FeaturesOE}.

	\subsection{Model Intervention}
	Model intervention encompasses techniques that go beyond identifying biases or errors in black-box models, allowing for direct adjustments that improve both accuracy and interpretability.
	For example, \citet{Moayeri2022HardIS} used attribution methods to identify spurious correlations in models trained on Hard-ImageNet and applied foreground constraints to refine model attention. Yet, attribution methods typically indicate only ``{\it where}" the model focuses, without explaining ``{\it what}" specific features are influential. As a result, explanations are often limited to the foreground of the object, preventing fine-grained attribution to critical features.

	In some cases, biases stem from inherent object characteristics \cite{Torralba2011UnbiasedLA, Szegedy2013IntriguingPO}. For instance, models may confuse ``{\it horse}'' and ``{\it zebra}'' due to shape similarity, but emphasizing texture and color, like ``{\it black and white stripes}'', improves classification. Identifying biases, whether in foreground or background features, is crucial for implementing human-understandable interventions that refine model behavior.
	
	Post-hoc methods allow limited intervention, while CBM-based approaches can be refined to the point of concept. CB2M \cite{Steinmann2023LearningTI} employs a memory buffer mechanism to automate and generalize CBM interventions, requiring user input on only a small number of samples. However, its intervention remains sample-level and still relies on prior user intervention. CCGM \cite{Dominici2024CausalCG} structures the concept bottleneck using a causal concept graph, modeling causal relationships in concept-based reasoning. While it provides intervention methods to explain the model’s causal reasoning, it is limited to single-modal natural language concepts. \citet{Marcinkevics2024BeyondCB} used probe functions to map black-box hidden-layer features into a concept bottleneck, enabling interpretation through language-based concepts and associated scores. This approach combines concept editing with human prior knowledge to address spurious correlations, and thus correct black-box errors.

	\section{Methodology}\label{sec:m}
	
	\noindent The feature attribution or concept explanation provided by existing post-hoc methods is actually intended to allow people to understand how neural networks work. At the same time, making the neural network understand human knowledge and correct its errors is the intention of the human neural network mutual understanding framework.

\begin{figure*}[!ht]
		\vspace{-0.5 cm}
		\centering
		\includegraphics[scale=0.48]{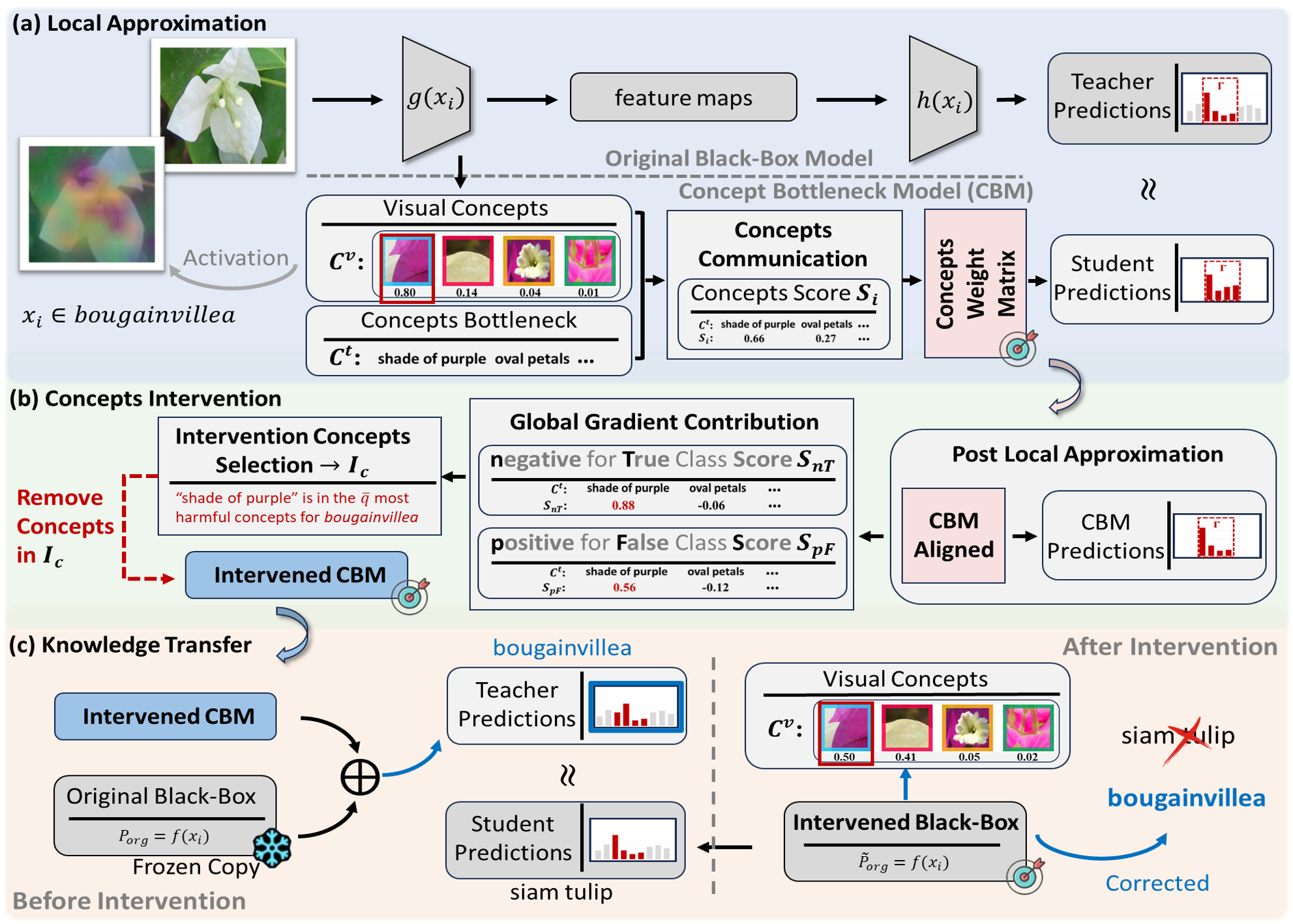}\vspace{-0.3cm}
		\caption{ \textbf{The Intervention Process}. \textbf{(a) Local approximation}: The CBM inference structure first distills an interpretable basis, that approximates the original black-box model’s class predictions on confused classes ($\Gamma$); 
			\textbf{(b) Concept intervention}: Next, the classification bias is corrected by intervening with concepts that are detrimental to the decision-making and selected based on the concept score of $S_{\text{nT}}$ and $S_{\text{pF}}$ in the CBM obtained from Step \textbf{(a)}; \textbf{(c) Knowledge transfer}:  Finally, the corrective adjustments made to the CBM in Step \textbf{(b)} are transferred back to the black-box model to improve reasoning and reduce biases. We refer to Figure \ref{fig:two} as an illustrative example to demonstrate intervention interpretation and predicted corrective effects.
		}
		\label{fig:one}
		\vspace{-0.5cm}
	\end{figure*}

	In this section, we formally define CBM-HNMU framework, illustrated in \textbf{Figure \ref{fig:one}}. We further explain how this framework facilitates mutual understanding between humans and neural networks, as well as how it enables the correction of erroneous predictions in black-box models.
	
	\subsection{Problem Setup}
	
	We first introduce the black-box classifier, denoted by the functional $f(x) = h\left(g(x)\right)$. Here, the feature activation map is denoted by $g(x): \mathbb{R}^{m\times m\times 3} \mapsto \mathbb{R}^{p}$,  where $p$ is the dimensionality of the feature representation and $m$ is the standard size (number of pixels) of input image $x$.
	The function $g(x)$ characterizes the feature extraction process in a hidden layer of the black-box classifier, while  $h(\cdot)$ serves as the classification head, mapping the extracted feature representation  $g(x)$  to the final logits.

	Additionally, we define the dataset used for training, validation, and inference  as the set of doublet of every image and its corresponding class, i.e.,
	$ \displaystyle
	D_{\text{train}} := \left\{(x, y)\right\}^{N_{\text{train}}}, 
	D_{\text{val}}:= \left\{(x, y)\right\}^{N_{\text{val}}}, 
	D_{\text{test}} := \left\{(x, y)\right\}^{N_{\text{test}}}   
	$
	where $N_{\text{train}}$, $N_{\text{val}}$, and $N_{\text{test}}$ denote the number of sample images in the training, validation, and test sets, respectively. Each sample consists of an image  $x$  and its corresponding class label  $y_{i}, ~\forall ~ i = 1, 2, \dots, N_{\text{class}}$, and $N_{\text{class}}$ represents the total number of classes.

	\subsection{Confused Classes Selection}
	
	We are now ready to investigate whether interpretable structures can approximate the reasoning processes of black-box models using a constrained set of concepts. However, constructing an inference model with a fixed, limited concept bottleneck that accurately captures the black-box model’s reasoning across all classes in a dataset remains a significant challenge. To this end, we begin by identifying specific classes within each dataset and model that are frequently misclassified as one another. We define confused samples as instances where the model erroneously predicts one similar class as another. Using the black-box classifier trained on  $D_{\text{train}}$, we evaluate its performance on $D_{\text{val}}$ and record the number of misclassified samples for each class pair.

	Next, we rank these class pairs based on the frequency of misclassification and extract the pairs exhibiting the highest confusion. Finally, we identify the set of confused classes, denoted as $\Gamma \subseteq \{1, 2, \dots, N_{\text{class}}\}$, whose corresponding cardinality is known as the number of confused classes, $|\Gamma| = N_{\Gamma} \geq 2$.\footnote{Confused set must contain at least two elements for confusion to occur.} These classes are considered for intervention and corrective adjustments in the black-box model.
	\black

	\subsection{Concepts Communication}
	Once we identify the classes requiring intervention, we can begin constructing a human-neural network mutual understanding framework for the confused classes, $\Gamma$.  It is essential to align their understanding to establish a communication bridge between humans and neural networks. Methods such as ACE, IBD, MOCE, and CRAFT, among others, provide extraction techniques that help us gain insights into the black-box model. Among them, CRAFT performs efficient unsupervised concept extraction based on non-negative matrix factorization (NMF) \cite{Green2023AlgorithmsFN} and provides feature attribution while avoiding additional annotations. Thus, in our subsequent experiments, we adopt CRAFT as a black-box-driven visual concept extraction method, where each concept is represented as $C^{v}_{k} \in \mathbb{R}^{m \times m \times 3}$ for  $k = 1,2, \dots, n$. Given an input image  $x_i$ from the training/testing/validation set, the corresponding visual concepts are extracted using CRAFT, resulting in
	\begin{align}\label{eq:Cv}
		C^{v}(i) &= \text{CRAFT}\Big(g(x_{i})\Big),
	\end{align}
	so that $C^{v}: \{C_1^{v}, C_2^{v}, \dots, C_{n}^{v}\}$, with $n$ denotes the number of visual concepts used for mapping concept bottleneck.

	The most intuitive and easily interpretable concepts for humans are expressed through natural language. By leveraging a concept bottleneck generated using ChatGPT-3.5-Turbo, we can efficiently construct a transparent reasoning framework. To design prompt templates for generating concept bottlenecks across all datasets in our experiments, we directly adopt the methodology introduced in LaBo \cite{Yang2022LanguageIA}. Consequently, we obtain a concept bottleneck $C^{t} \in \mathbb{C}$ for each dataset. For example, in the Flower-102 dataset, the bottleneck is represented as
	$
	C^{t} = \{\text{shade of purple}, \text{ovalpetals}, \dots\},
	$
	with $|C^{t}| = N_c$ the total number of natural language concepts.
	\black
	
	To efficiently establish connection between $C^v$ and $C^t$, CBM-HNMU utilizes CLIP as an intermediary. The visual concepts extracted from the black-box model’s intermediate layer and the natural language concepts in the concept bottleneck are mapped into a shared representation space  $\mathbb{R}^{1 \times d}$ ($d = 512$). This transformation is performed using CLIP’s image encoder $E_{\text{img}}(C^{v}_{k}): \mathbb{R}^{m \times m \times 3} \mapsto \mathbb{R}^{1 \times d}$ for $k = 1,2,\dots,n$, and its text encoder  $E_{\text{text}}(C^{t}): \mathbb{C} \mapsto \mathbb{R}^{N_{c} \times d}$.

	Subsequently, concept score $S_{i} \in \mathbb{R}^{1 \times N_{c}}$ is computed for each input image $x_i \in D_{\text{train},\text{val},\text{test}}$  by  
	\begin{align}\label{eq:1}\nonumber
		S_i = & \text{ CS} (g(x_{i}), C^{t})  = \frac{1}{n} \sum_{k=1}^{n} E_{\text{img}}(C_k^{v}(i)) \times E_{\text{text}}(C^{t})^{\mathrm{T}} \\
		= &\frac{1}{n} \sum_{k=1}^{n} E_{\text{img}}(\text{CRAFT}(g(x_{i}))_{k}) \times E_{\text{text}}(C^{t})^{\mathrm{T}},
	\end{align}
	where $\times$ denotes the matrix multiplication and $\text{CS}(\cdot)$ the function representing the acquisition process of the concept score $S_i$. The number of visual concepts used to map the concept bottleneck is specified by $n$. The notation $\text{CRAFT}(g(x_{i}))_{k} = C_k^{v}(i)$ refers to the $k$-th visual concept extracted from the hidden layer of the black-box model. The computed concept score $S_i$ provides a unified representation of the model’s interpretation of an input image $x_i$ in terms of human-understandable concepts. From the score distribution across natural language concepts, humans can gain insight into how the black-box model internally processes and understands different aspects of the image.

	\subsection{Local Approximation}
	
	As illustrated in \textbf{Figure \ref{fig:one}.(a)}, we first obtain the concept score vector $S_i$, which is interpretable for both humans and black-box models. To construct the inference component of CBM, we use a concept weight matrix  $W \in \mathbb{R}^{N_{\Gamma} \times N_c}$. This matrix is used to build a classification header similar to that of the original model.
	
	For each input image $x_i$, the corresponding concept score vector $S_i$ is passed through CBM, producing the classification output $P_{\text{cbm}}: \mathbb{R}^{m\times m\times 3} \mapsto \mathbb{R}^{1 \times N_{\Gamma}}$ as 
	\begin{align}\label{eq:cbm}
		P_{\text{cbm}}(x_i) =  S_i \times W^{\mathrm{T}} = \text{CS} (g(x_i), C^{t}) \times W^{\mathrm{T}}.
	\end{align}
	Simultaneously, the original black-box model generates its own predictions  $P_{\text{org}}: \mathbb{R}^{m\times m\times 3} \mapsto \mathbb{R}^{1 \times N_{\text{class}}}$ that reads
	\begin{align}
		P_{\text{org}}(x_i) = f(x_i) = h(g(x_i)).    
	\end{align}
	Next, we extract the original black-box predictions $P_{\text{org}}^{\Gamma} \in \mathbb{R}^{1 \times N_{\Gamma}}$ from confused classes $\Gamma$, and align them with CBM’s output using the $\ell_2$-norm. The local approximation loss function is then defined as
	\begin{align}
		\text{Loss}_{\text{lp}} = \frac{1}{N_{\Gamma}} \|P_{\text{cbm}}(x_i) - P_{\text{org}}^{\Gamma}(x_i)\|_2,    
	\end{align}
	where $P_{\text{org}}^{\Gamma}$ represents the subset of the black-box predictions corresponding to the confused classes.
	
	Our goal is to reduce the bias and complexity of model approximation by utilizing similar concept expressions and structures. This allows the CBM to take input aligned with the visual concepts extracted from the black-box's hidden layer, and make predictions through a simple linear layer that mimics the black-box behavior.

	% \blue 
	% \newpage
	\subsection{Concepts Intervention}

	\citet{Lewis1994HeterogeneousUS}, \citet{Esteva2017DermatologistlevelCO}, \citet{Yuan2021MultipleIA}, \citet{Guo2007OptimisticAU}, among others, have provided valuable insights into misleading concept selection and modification. As shown in \textbf{Figure \ref{fig:one}.(b)}, our intervention method is based on gradient contribution.  The potential concept errors within locally confused classifications may fall into the following two categories
	\begin{enumerate}
		\item[i)] Concept $C_j^t$ with \textbf{negative} impact on the correct (\textbf{True}) class predicted by CBM, that quantified by $S_{\text{nT}}$.
		\item[ii)] Concept $C_j^t$ with \textbf{positive} influence on CBM's incorrect (\textbf{False}) prediction of other classes, quantified by $S_{\text{pF}}$.
	\end{enumerate}
	These two errors are represented by concept scores and quantified using contribution matrices $S_{\text{nT}}, S_{\text{pF}} \in \mathbb{R}^{N_{\Gamma} \times N_{c}}$, which capture the statistical impact of each concept on global model behavior for each confusion class.

	% \blue
	To construct the concept contribution matrices $S_{\text{nT}}$ and $S_{\text{pF}}$, consider a sample $x_i \in D$ with ground-truth label $y_i$. Let $y^*_{i}:= \displaystyle \max_j [P_{\text{cbm}}(x_i)]_{1,j}$ denotes predicted class under the concept bottleneck model and $W \in \mathbb{R}^{N_\Gamma \times N_c}$ the concept weight matrix, where each row vector $w_k \in \mathbb{R}^{1 \times N_c}$ corresponds to confusion class $k$.
	
	To assess the influence of concepts on model behavior, we define a gradient-based attribution function
	\begin{equation}
		G(w_k, P_k(x_i)) := \frac{\partial P_k(x_i)}{\partial w_k} \odot w_k,    
	\end{equation}
	where $\odot$ denotes element-wise multiplication and $P_k(x_i) = [P_{\mathrm{cbm}}(x_i)]_{1,k}$ is the predicted probability for class $k$. Based on $S_i \in \mathbb{R}^{1 \times N_c}$ from Eq.~\eqref{eq:1}, the per-sample contribution scores are defined as follows:
	\begin{subequations}
		\begin{align}\label{eq:concmatrix1}
			S_{\text{nT}}(i)_{y_i} &:=
			\begin{cases}
				-S_{i} \odot G(w_{y_i},P_{y_{i}}(x_{i})), \;\; & \text{ if } y_{i} \in \Gamma \\
				\phantom{-} \mathbf 0, \;\;  & \text{ otherwise}
			\end{cases}\\ \label{eq:concmatrix2}
			S_{\text{pF}}(i)_{y^*_{i}} &:=
			\begin{cases}
				\phantom{-}S_{i} \odot G(w_{y_i^*},P_{y^*_{i}}(x_{i})),  \;\;  & \text{ if } y^*_{i} \neq y_{i} \\
				\phantom{-} \mathbf 0, \;\;  & \text{ otherwise}
			\end{cases}.
			%\vspace{-0.3in}
		\end{align}  
	\end{subequations}
	Here, $S_{\text{nT}}$ captures how the activated concepts suppress the true class, while $S_{\text{pF}}$ measures how they reinforce the incorrect predicted class. To mitigate the impact of unfavorable concepts, we accumulate the sample-wise scores over the validation set to form global contribution matrices. Concepts in each confused class are ranked by their total influence in $S_{\text{nT}}$ and $S_{\text{pF}}$, and the top $\overline{q}$ most detrimental concepts are selected for removal during intervention.

	We summarize the concept intervention procedure in Algorithm~\ref{a:1}, which identifies and prunes detrimental concepts based on their contribution to prediction errors.
	\begin{algorithm}[htb!]
		\caption{Gradient-Based Concept Intervention}
		\renewcommand{\algorithmicrequire}{\textbf{Input:}}
		\begin{algorithmic}[1]
			\REQUIRE Validation set $D_{\text{val}}$, concept weight matrix $W$, encoders $E_{\text{img}}$, $E_{\text{text}}$
			\STATE Initialize contribution matrices $S_{\text{nT}} \leftarrow 0$, $S_{\text{pF}} \leftarrow 0$
			\FOR{$i = 1$ to $N_{\text{val}}$}
			\STATE Compute $S_i$, $P_{\text{cbm}}(x_i)$, and predicted class $y_i^\star$
			\STATE Update $S_{\text{nT}}(i)_{y_i}$ and $S_{\text{pF}}(i)_{y_i^\star}$ using Eqs.~\eqref{eq:concmatrix1} and \eqref{eq:concmatrix2}, then accumulate $S_{\text{nT}}$ and $S_{\text{pF}}$
			\ENDFOR
			\FOR{each confused class $k \in \Gamma$}
			\STATE Select top $q/2$ concepts $I_{\text{nT}}^{k}>0$ from $S_{\text{nT}}^k$.
			\STATE Select top $q/2$ concepts $I_{\text{pF}}^{k}>0$ from $S_{\text{pF}}^k$.
			\STATE Merge selected concepts as $I_c^{k} = \{I_{\text{nT}}^{k}\} \bigcup \{I_{\text{pF}}^{k}\}$.
			\ENDFOR
			\RETURN Intervention indices $I_c \in \mathbb{R}^{N_{\Gamma} \times \bar{q}}$ with $\bar{q} \leq q$
		\end{algorithmic}
		\label{a:1}
	\end{algorithm}
	
	After identifying the intervention index set $I_c$ from $D_{\text{val}}$, we update the concept weight matrix $W \in \mathbb{R}^{N_{\Gamma} \times N_c}$ by zeroing out the selected concept weights for each confused class. Specifically, for each $k \in \Gamma$ and each concept $j \in I_c^k$, we set $W[k,j] = 0$. The resulting $\overline{W}$ defines the intervened concept weight matrix of the CBM.

	\subsection{Knowledge Transfer}
	
	\noindent 
	After the intervened CBM ($\overline{W}$) is updated, the next step is to transfer the refined knowledge back into the original black-box model. As illustrated in \textbf{Figure \ref{fig:one}.(c)}, since CBM provides a localized approximation of the black-box model, we retain the learned knowledge for all classes except those involved in confusion. To achieve this, we construct a teacher $P_t \in \mathbb{R}^{1 \times N_{\text{class}}}$  by combining two sources  of predictions 
	\begin{enumerate}
		\item[i)] Frozen copy of the original black-box predictions for the unaffected classes denoted as $P_{\text{org}}^{\Gamma^{\complement}} \in \mathbb{R}^{1 \times N_{\Gamma^{\complement}}}$, where $\Gamma^{\complement}:= \{1, \dots, N_{\text{class}}\} \setminus \Gamma$ and $N_{\Gamma^{\complement}} = N_{\text{class}} - N_{\Gamma}$.
		\item[ii)] Post-intervention CBM predictions $P^{\star}_{\text{cbm}} \in \mathbb{R}^{1 \times N_{\Gamma}}$, that refine model’s reasoning for the confused classes.
	\end{enumerate}
	The combined output $P_t$ serves as a supervisory signal for reverse distillation, guiding the original black-box model $P_s \in \mathbb{R}^{1 \times N_{\text{class}}}$ to integrate the improvements introduced by CBM while preserving its original predictions for non-confused classes. Given the distillation temperature $T_1$ of $P^{\star}_{\text{cbm}}$, the teacher’s prediction reads
	\begin{align} \nonumber
		P_t &= \mbox{combine} (P^{\star}_{\text{cbm}}, P^{\Gamma^{\complement}}_{\text{org}}) \\ 
		&= \mathrm{pr} \cdot \mathrm{softmax} (P^{\star}_{\text{cbm}}/T_1) \oplus \mathrm{softmax}(P_{\text{org}})^{\Gamma^{\complement}},    
	\end{align}
	where $\mathrm{softmax}(P_{\text{org}})^{\Gamma^{\complement}}$ and $P^{\star}_{\text{cbm}}$ represents the predicted probability of the confusion classes by the original black-box model with frozen parameters, and the logits output of the post-intervention CBM, receptively. The operator $\oplus$ denotes the concatenation of output elements.

	To ensure that the model’s predictions for non-intervention classes remain unchanged, we introduce the probability residual coefficient $\mathrm{pr}$, defined as
	\begin{align}
		\mathrm{pr} = 1 - | \mathrm{softmax}(P_{\text{org}})^{\Gamma^{\complement}} |_1,
	\end{align}
	whereby the residual probabilities of the intervention classes from the original black-box model onto the post-intervention CBM are redistributed. 
	
	The student’s prediction  $P_s$ is given by 
	\begin{equation}
		P_s = \mathrm{softmax}(\tilde{P}_{\text{org}}/T_2),   
	\end{equation}
	where $\tilde{P}_{\text{org}}$ represents the black-box predictions for the subset of parameters requiring intervention, and $T_2$ is the distillation temperature for the original black-box model.
	
	Finally, we perform black-box intervention by minimizing the cross-entropy loss between $P_t$ and $P_s$, i.e.,
	\begin{equation}
		\mathrm{Loss}_{\text{kt}} = \sum^{N_{\text{class}}}_{k=1} P_t^k \cdot \log(P_s^k),    
	\end{equation}
	over the validation dataset $D_{\text{val}}$. This ensures that refined knowledge from CBM is effectively transferred back to the black-box model.

	\section{Experiments}

	\subsection{Experiments Setups}
	\begin{table*}[ht]
		\centering
		\scalebox{0.70}{
			\begin{tabular}{cccccccccccc}
				\hline
				\multirow{2}{*}{Models} & \multicolumn{2}{c}{Flower-102} & \multicolumn{2}{c}{CIFAR-10} & \multicolumn{2}{c}{CIFAR-100} & \multicolumn{2}{c}{CUB-200} & \multicolumn{2}{c}{FGVC-Aircraft} & \multirow{2}{*}{AVG IMP}\\
				\cline{2-11}
				& $w/o$ INT & $w/$ INT  & $w/o$ INT & $w/$ INT  & $w/o$ INT & $w/$ INT  & $w/o$ INT & $w/$ INT  & $w/o$ INT & $w/$ INT & \\
				\hline
				NFResNet50 & 94.28 & \textbf{95.36 $\pm{0.019}$}  & 80.92 & \textbf{83.56 $\pm{0.005}$} & 73.58 & \textbf{73.92 $\pm{0.037}$} & 62.41 & \textbf{62.67 $\pm{0.057}$} & 64.63 & \textbf{64.94 $\pm{0.028}$} & $\uparrow \textbf{0.93}$ \\
				Vit\_Small & 97.24 & \textbf{98.20 $\pm{0.019}$} & 91.51 & \textbf{92.84 $\pm{0.154}$} & 81.26 & \textbf{82.43 $\pm{0.031}$} & 74.75 & \textbf{75.39 $\pm{0.029}$} & 69.88 & \textbf{70.94 $\pm{0.251}$} & $\uparrow \textbf{1.03}$ \\
				GCVit & 93.58 & \textbf{95.20 $\pm{0.050}$} & 80.20 & \textbf{80.97 $\pm{0.054}$} & 72.38 & \textbf{72.55 $\pm{0.034}$} & 76.99 &\textbf{77.64 $\pm{0.070}$} & 69.81 & \textbf{71.41 $\pm{0.171}$} & $\uparrow \textbf{0.96}$ \\
				\hline
		\end{tabular}}
		\caption{Comparison of global classification accuracy without ($w/o$) or with ($w/$) intervention (INT). All experimental models using CBM-HNMU for black-box intervention have objective generalization performance improvements on the $D_{\text{test}}$ compared with the baseline. Results are presented as average $\pm$ standard deviation (AVG $\pm$ STD) over three random tests. The average improvement (AVG IMP) represents the mean accuracy increase for each row.
		}
		\label{tab:1}
		\vnsp
		\vspace{-0.3 cm}
	\end{table*}
	
	\begin{table*}[hbp]
		\centering
		\scalebox{0.85}{
			\begin{tabular}{cccccccc}
				\hline
				\multirow{2}{*}{Models} & \multicolumn{2}{c}{Flower-102} & \multicolumn{2}{c}{CUB-200} & \multicolumn{2}{c}{FGVC-Aircraft} & \multirow{2}{*}{AVG IMP}\\
				\cline{2-7}
				& $w/o$ INT & $w/$ INT  & $w/o$ INT & $w/$ INT & $w/o$ INT & $w/$ INT & \\
				\hline
				ResNeXt26 & 88.47 & \textbf{90.85 $\pm{0.363}$} & 59.32 & \textbf{59.60 $\pm{0.113}$} & 64.27 & \textbf{64.72 $\pm{0.212}$} & $\uparrow \textbf{1.04}$ \\
				BotNet26 & 89.20 & \textbf{91.30 $\pm{0.050}$}  & 56.89 &  \textbf{57.09 $\pm{0.066}$} & 53.70 & \textbf{54.51 $\pm{0.079}$} & $\uparrow \textbf{1.04}$ \\
				RexNet100 & 90.26 & \textbf{90.77 $\pm{0.082}$} & 55.40 & \textbf{57.39 $\pm{0.024}$} & 52.60 & \textbf{52.98 $\pm{0.079}$} & $\uparrow \textbf{0.96}$ \\
				\hline
			\end{tabular}
		}
		\caption{
			Global classification accuracy with (w/) and without (w/o) intervention (INT) of another three architectures. Results are reported as the average $\pm$ standard deviation (AVG $\pm$ STD) over three random tests, and AVG IMP as the average improvement for each row.
		}
		\label{tab:1.1}
		\vnsp
	\end{table*}
	
	\begin{figure*}[!htp]
		\vspace{-0.3 cm}
		\centering
		\includegraphics[scale=0.475]{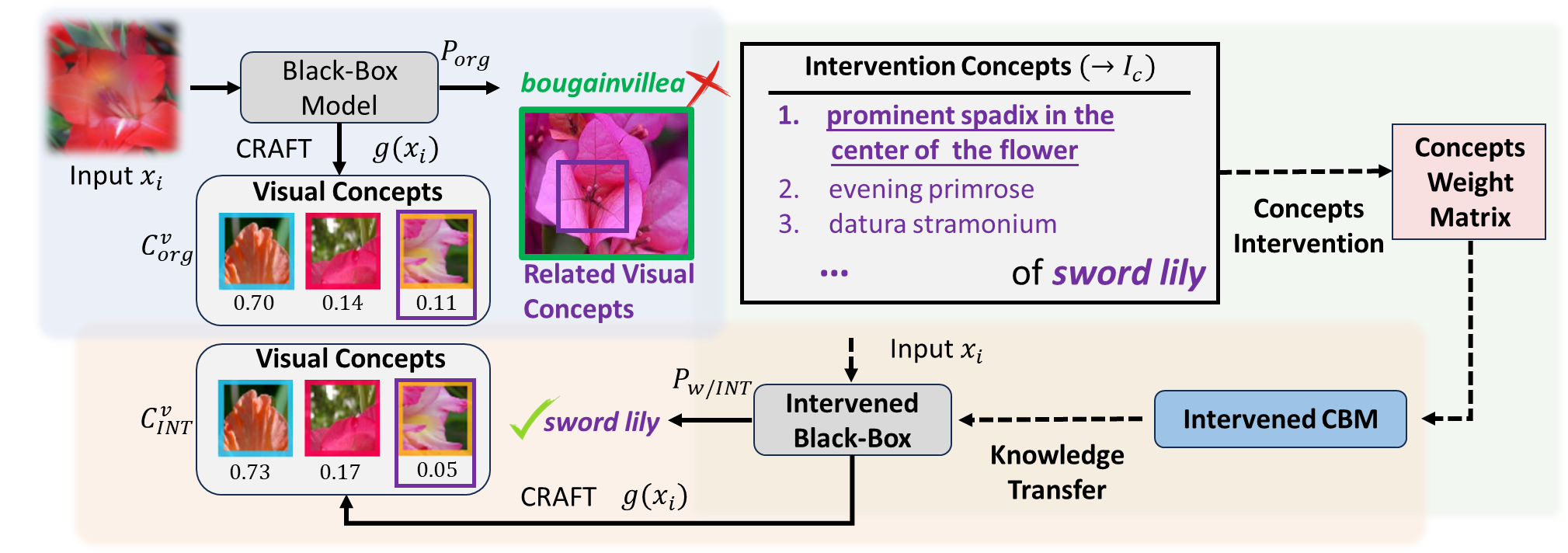}
		\caption{ {\bf Comparative visualization of intervention concepts and black-box attributions before and after intervention.} It shows the visual results of NFResNet50 on Flower-102. Through {\bf concept intervention}, CBM-HNMU identify and remove $\overline{q}$ harmful concept ($I_{c}$) from confusing class (\textcolor{purple}{\bf``sword lily"}) based on $S_{\text{nT}}$ and $S_{\text{pF}}$. After intervention knowledge is transferred back through CBM, the black-box model corrects the misclassified \textcolor{dgreen}{\bf ``bougainvillea"}. Analyzing the visual concept changes in the black-box model’s middle layer before and after intervention, we observe a decrease in the ``total Sobol indices" of the \textcolor{purple}{Related Visual Concept}, which is corresponding to \textcolor{purple}{\bf``prominent spadix in the center of the flower"}. Compared with the bougainvillea sample marked by green border, we find that the flower center features of both classes are similar, which likely contributed to the misclassification. The intervention reduces the model’s reliance on such subtle distinctions, leading to more accurate predictions. For additional visual explanations, see \textcolor{blue}{\bf Appendix. H}.}
		\label{fig:two}
		\vnsp
		\vspace{-0.3 cm}
	\end{figure*}
	
	\noindent \textbf{Dataset.}  
	CBM-HNMU is evaluated on Flower-102 \cite{Nilsback2008AutomatedFC}, CIFAR-10, CIFAR-100 \cite{krizhevsky2009learning}, FGVC-Aircraft \cite{Maji2013FineGrainedVC}, and CUB-200 \cite{wah2011caltech}. CIFAR-10 and CIFAR-100 represent object datasets of varying scales with high discriminative power, while Flower-102, CUB-200, and FGVC-Aircraft contain objects with fewer distinguishing features, necessitating more fine-grained classification.

	\noindent \textbf{Baselines.} We evaluate CBM-HNMU across classic and commonly used CNN and Transformer architectures, including NFResNet50 \cite{Mittal2022NFResNetMA}, ResNeXt26 \cite{Xie2016AggregatedRT}, BotNet26 \cite{Srinivas2021BottleneckTF}, RexNet100 \cite{Han2020ReXNetDR}, ViT-Small \cite{Dosovitskiy2020AnII}, GCVit \cite{Hatamizadeh2022GlobalCV}, CaiT-Small \cite{Touvron2021GoingDW}, ConVit-Base \cite{d2021convit}, and DeiT-Base \cite{Touvron2020TrainingDI}.

	\noindent \textbf{Evaluation Protocol.} All baselines are initialized with ImageNet-1K \cite{imagenet15russakovsky} pre-trained weights and trained for 50 epochs on different $D_{\text{train}}$. Confusion class selection, local approximation, concept intervention, and intervention knowledge transfer are performed on validation set $D_{\text{val}}$, while evaluations before and after black-box intervention are conducted on testing set $D_{\text{test}}$. The local approximation step is with learning rate of $1e^{-4}$ and runs for $200$ epochs, and intervention knowledge transfer is trained with a learning rate of $3e^{-7}$ for $10$ epochs.
	
	\subsection{Experiments Results} 
	We evaluate the improvements in both global accuracy and fine-grained class prediction, resulting from the CBM-HNMU intervention across multiple datasets. In addition, we examine how the visual concepts captured by black-box model evolve before and after the intervention, using visualization techniques and natural language concepts to interpret the process. An ablation study further illustrates how the quality of black-box approximation and the overall intervention effect vary with the number of intervention concepts.
	% Herein, we compare global accuracy and fine-grained class prediction improvements brought by the CBM-HNMU intervention from various perspectives on multiple datasets. In addition, we comparatively analyze the visual concepts changes of the black-box model before/after the intervention based on visualization and explain the intervention process in combination with natural language concepts. Ablation study provides the trend of black-box approximation and intervention effect with the number of intervention concepts. 
	Furthermore, we examine the impact of CBM-based classification methods (CBM and Post-hoc CBM) and the influence of the number of confused classes ($N_{\Gamma}$) on intervention effectiveness. Finally, we conducted a user study to confirm that the chosen intervention concepts (i) align with human visual perception and (ii) help correct classification errors.

	\noindent \textbf{Performance Improvements.} As shown in \textbf{Table \ref{tab:1}}, the three architectures in our experiments achieved global accuracy improvements across five datasets after intervention with CBM-HNMU. To further verify the effect of CBM-HNMU, we conduct the identical comparative test using another three architectures on Flower-102, FGVC-Aircraft, and CUB-200. The results are presented in \textbf{Table \ref{tab:1.1}}. 
	As a result, CBM-HNMU achieved the highest accuracy improvement with NFResNet50 on CIFAR-10, increasing from $80.92\%$ to $83.56\%$, a gain of $2.64\%$. Additionally, Vit\_Small achieved the highest average improvement (AVG IMP) across the five datasets, with an increase of $1.03\%$.
	
	% \textbf{Table \ref{tab:2}} shows the number of corrected error samples and the coverage of confusion classes for different intervention models on $D_{\text{test}}$ of Flower-102. The notation ``$-n$'' (left of the bracket) indicates that $n$ samples are originally misclassified into a confusion class but are corrected after intervention. Conversely, ``$+n$'' (right of the bracket) represents $n$ samples that are initially misclassified into any class but are reassigned to a specific confusion class post-intervention. We observed that most corrected samples in $D_{\text{test}}$ are associated with intervened classes -- either originally misclassified as one of the confused classes or having a true label within the confused set but predicted as another class. Additionally, the classification accuracy of most non-confused classes remains unchanged after intervention. It can be observed that all black-box models corrected generalization errors, and thus improve classification performance.
	\textbf{Table \ref{tab:2}} shows the number of corrected samples and the coverage of confusion classes in different intervention models on $D_{\text{test}}$ of Flower-102. The notation ``$-n$'' indicates that $n$ samples originally misclassified into a confusion class are corrected after intervention. Conversely, ``$+n$'' denotes $n$ samples initially misclassified into any class that are reassigned to the confusion class after intervention. Most corrected samples in $D_{\text{test}}$ are related to the intervened classes, either because they were previously predicted as one of the confusion classes or because their true labels belong to the confusion set.
	% \textbf{Table \ref{tab:2}} shows the number of corrected samples and the coverage of confusion classes for different intervention models on $D_{\text{test}}$ of Flower-102. The notation ``$-n$'' (left of the bracket) indicates that $n$ samples are originally misclassified into a confusion class but are corrected after intervention. Conversely, ``$+n$'' (right of the bracket) represents $n$ samples that are initially misclassified into any class but are reassigned to a specific confusion class post-intervention. We observed that most corrected samples in $D_{\text{test}}$ are associated with intervened classes -- either originally misclassified as one of the confused classes or having a true label within the confused set but predicted as another class.
	
	\begin{table}[ht]
		%\vspace{-0.3 cm}
		\centering
		\scalebox{0.68}{
			\begin{tabular}{l>{\raggedright\arraybackslash}p{6cm}cc}
				\hline
				\multirow{2}{*}{Models} & \multicolumn{3}{c}{Flower-102 ($D_{\text{test}}$)} \\
				\cline{2-4}
				& Confusion Class ($\Gamma$)  & Corrected & Coverage\\ 
				\hline
				\multirow{4}{*}{NFResNet50} & \multirow{4}{*}{\makecell[l]{bougainvillea ($-6$,$+0$),\\ camellia ($-1$,$+2$),  mallow (-), \\ gaura ($-0$,$+2$), cyclamen (-), \\ sweet pea ($-9$,$+0$), sword lily ($-0$,$+5$)}} & \multirow{4}{*}{34} & \multirow{4}{*}{25} \\
				& & & \\
				& & & \\
				& & & \\
				\hline
				\multirow{3}{*}{BotNet26} & \multirow{3}{*}{\makecell[l]{bougainvillea ($-0$,$+3$),\\ camellia ($-22$,$+0$), hibiscus ($-12$,$+0$),\\ mallow ($-0$,$+1$), petunia ($-0$,$+1$)}} & \multirow{3}{*}{59} & \multirow{3}{*}{39} \\
				& & & \\
				& & & \\
				\hline
				\multirow{3}{*}{GCVit} & \multirow{3}{*}{\makecell[l]{petunia ($-18$,$+0$),\\ camellia ($-6$,$+0$),  mallow (-),\\ hibiscus ($-4$,$+0$), morning glory (-)}} & \multirow{3}{*}{51} & \multirow{3}{*}{28} \\
				& & & \\
				& & & \\
				\hline
		\end{tabular}}
		\caption{Statistics of fine-grained class interventions in Flower-102. ``Corrected'' refers to the total number of error samples corrected by each model on $D_{\text{test}}$, and ``Coverage'' represents the number of corrected samples associated with the confused classes.
		}
		\label{tab:2}
		\vspace{-0.3 cm}
	\end{table}
	
	\noindent \textbf{Intervention Visualization.}  Herein, We analyze how the intervened black box’s performance improvement relates to changes in natural language and visual-related concepts. \textbf{Figure \ref{fig:two}} presents comparative results illustrating the intervention process on Flower-102 dataset, highlighting visual concept explanations before and after intervention. Samples are randomly selected from a subset, where the black-box model’s original classification errors on the test set are corrected following intervention. Each corrected sample includes at least one pre- or post-correction class related to the confused classes, ensuring a clear visual link to the intervention concept (refer to ``Coverage" in \textbf{Table \ref{tab:2}}).
	
	We observe that intervened natural language concepts often shift black-box visual concept explanations. In the visualization, the black-box model misclassifies sword lily as bougainvillea, relying on feature ``prominent spadix in the center of the flower,'' that shared by both classes. The corresponding concept score drops from $0.11$ to $0.05$ after intervention, indicating reduced reliance on this feature. CBM-HNMU refocuses the model from the spadix to petal-related features, demonstrating that it does not merely transfer knowledge, but actively corrects black-box errors by aligning human and neural network reasoning.
	\begin{table*}[ht]
		\centering
		\scalebox{0.9}{
			\begin{tabular}{ccccccc}
				\hline
				\multirow{2}{*}{Models} & \multicolumn{2}{c}{Flower-102} & \multicolumn{2}{c}{CUB-200} & \multicolumn{2}{c}{FGVC-Aircraft} \\
				\cline{2-7}
				& w/ OURS & w/ RAND  & w/ OURS & w/ RAND  & w/ OURS & w/ RAND \\
				\hline
				NFResNet50 & 95.36 & 94.72 ($\downarrow 0.64$)  & 62.67 & 62.51 ($\downarrow 0.16$) & 64.94 & 64.75 ($\downarrow 0.19$) \\
				Vit\_Small & 98.20 & 97.81 ($\downarrow 0.39$) & 75.39 & 75.23 ($\downarrow 0.16$) & 70.94 & 69.97 ($\downarrow 0.97$) \\
				\hline
			\end{tabular}
		}
		\caption{Comparison of global classification accuracy using CBM-HNMU with intervention method (w/OURs) and randomized concept intervention (w/RAND). Black-box model optimized by CBM-HNMU outperforms random intervention under the same parameter settings.}
		\label{tab:3}
		\vnsp
	\end{table*}
	
	% 这个实验在附加材料详细提了，这里就不放了，而且和用户实验想说明的观点是基本一致的（干预概念符合人类视觉认知，以及具有混淆类别误导特性）
	% \noindent \textbf{Intervention Concept Visual Masking.} 
	% To further investigate the relationship between natural language concepts removed during intervention and the erroneous visual dependencies of the original black box, we represent the corresponding vision-related intervention concepts as masks and apply them to misclassified samples. From hypothesis, masking visual features associated with intervening concepts should improve the black box’s classification performance on these erroneous samples. As in \textbf{Figure \ref{fig:three}}, it shows that masking generally leads to error correction or an increase in the black box’s confidence in the correct class.

	% \begin{figure}[htb!]
		% \centering
		% \includegraphics[width=\linewidth]{figure3_own.png}
		% \caption{ Intervention concept visual masking results across models and datasets. Different confusion classes are color-coded along with their corresponding visually related intervened concepts. Notably, in the third row, since intervention concept of ``Common Tern" appears ``gray", a white mask is used to avoid impact.
			% }
		% \label{fig:three}
		% \end{figure}

\begin{figure}[htp]
		%\vspace{-0.3 cm}
		\centering
		\includegraphics[trim={0cm 0.1cm 0cm 0.8cm},clip,width=0.9\linewidth]{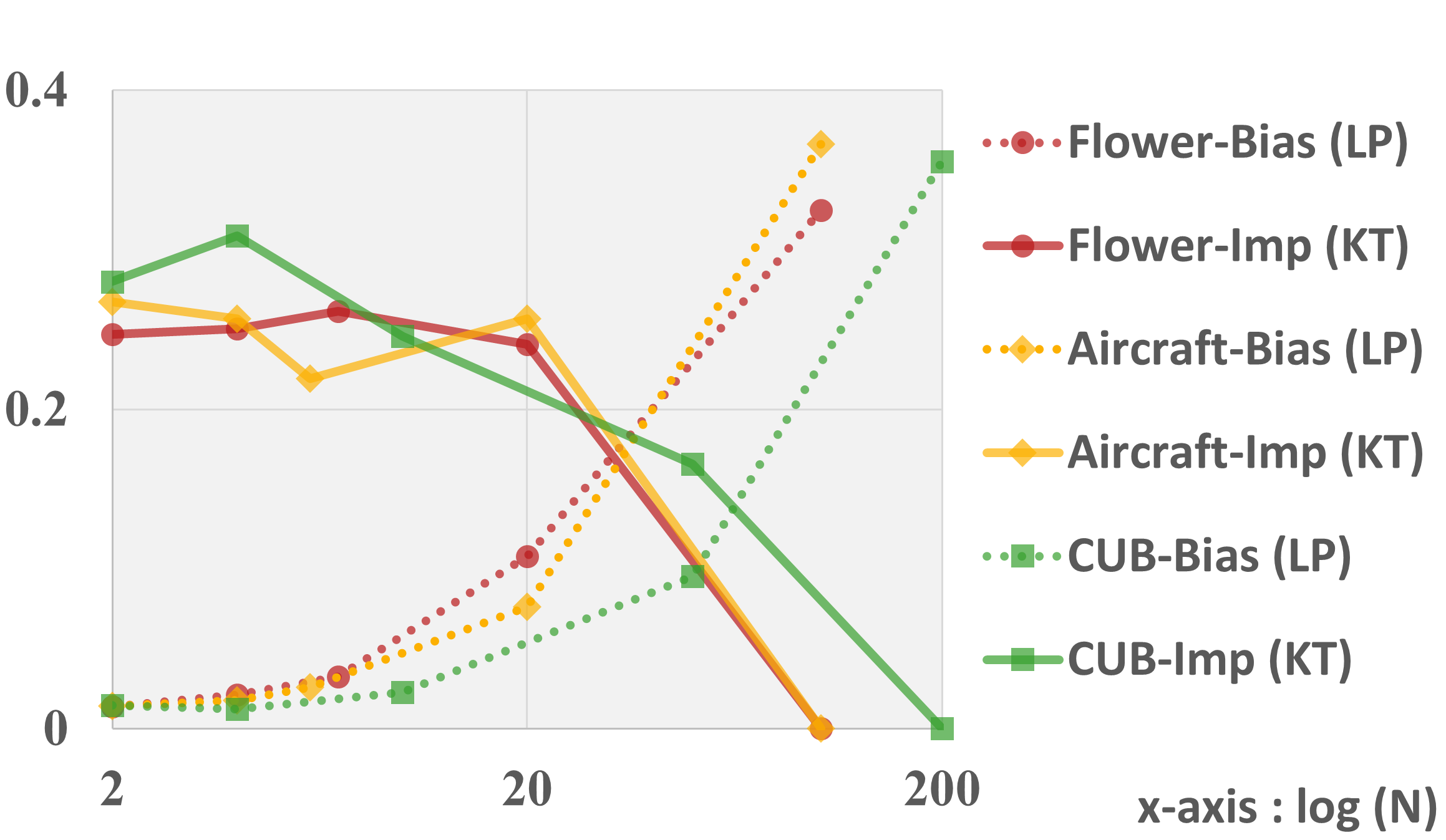}
		\caption{
			The dotted line represents the normalized category prediction bias of the local approximate (LP) CBM and the original NFResNet-50 model across three datasets ($D_{\text{test}}$) under different numbers of intervention classes. The solid line shows the normalized accuracy improvement (Imp) after intervention knowledge transfer (KT) back to the original model. 
		}
		\label{fig:four}
		\vnsp
		\vspace{-0.3 cm}
	\end{figure}

	\noindent \textbf{Ablation Study.} 
	In \textbf{Figure \ref{fig:four}}, after ablating the number of confused classes across the three datasets, when the proportion of intervention classes is less than $25\%$, local approximation bias and the intervention improvement are relatively stable. When all classes are intervened, approximation bias and intervention improvement deteriorate significantly.
	
	To further validate the effectiveness of our intervention concept selection method, we conducted a random concept intervention for reverse knowledge transfer, applying CBM-HNMU on Flower-102, CUB-200, and FGVC-Aircraft using NFResNet50 and ViT. The results in \textbf{Table \ref{tab:3}}, show that randomized intervention (often used as a baseline for CBM intervention methods) performs poorly. Our findings align with this expectation, showing that CBM-HNMU achieves an average accuracy improvement of $0.42$ over random intervention, with the highest improvement of $0.97$.
	
	% 概念替换这里就不提了，附加材料详细说了
	% Finally, since deleting concepts alone may not correct errors stemming from concept bottleneck limitations, we introduce a concept replacement method using an additional concept search set (see {\color{blue} \bf Appendix. B}).

	\noindent \textbf{Performance Comparison with CBM-Related Methods.} 
	We evaluate the performance of the black-box model after CBM-HNMU intervention using NFResNet-50 and compare it with CBM (same backbone and concept bottleneck) and Post-hoc CBM across three datasets. As shown in \textbf{Table \ref{tab:2.1}}, CBM-HNMU achieves best classification performance. Additionally, we observe that Post-hoc CBM improves performance over CBM by incorporating the backbone’s residual expression. However, it still incurs accuracy loss compared to the corresponding black box model.
	
	\begin{table}[htb!]
		%\vspace{-0.2 cm}
		\centering
		\scalebox{0.85}{
			\begin{tabular}{c|ccc}
				\hline
				\diagbox{Models}{Dataset} & Flower-102 & CUB-200 & FGVC-Aircraft \\
				\hline
				Black-Box & 94.28 & 62.41 & 64.63 \\
				\rowcolor{gray!20} CBM & 92.57 & 54.53  & 56.31 \\
				\rowcolor{gray!20} PCBM & 93.58 & 58.20 & 64.45 \\
				\rowcolor{gray!100} \textcolor{white}{CBM-HNMU} & \textcolor{white}{\textbf{95.36}} & \textcolor{white}{\textbf{62.67}} & \textcolor{white}{\textbf{64.94}} \\
				\hline
			\end{tabular}
		}
		\caption{Models optimized with CBM-HNMU intervention show significant improvements over corresponding baseline, CBM, and Post-hoc CBM across three datasets (NFResNet50 as backbone).}
		\label{tab:2.1}
		\vnsp
		%\vspace{-0.1 cm}
	\end{table}
	
	%\noindent \textbf{User Study.} 
	%Our model reduces reliance on misleading concepts tied to fine-grained confusion, leading to minor changes in attribution maps but notable shifts in concept importance scores. To validate the selected intervention concepts align with human perception and aid in correcting classification errors, we conducted a user study on Flower-102, CUB-200, and FGVC-Aircraft. For each confused class in the test set, we clustered samples into ten groups and selected one representative per cluster as reference examples for participants. We recruited $30$ paid, non-expert volunteers to assess each intervention concept according to whether it matched true class (CT), false class (CF), or semantically rational (CR), as shown in \textbf{Figure} \ref{fig:five}.
	
\begin{figure}[!ht]
		%\vspace{-0.35 cm}
		\centering
		\includegraphics[width=\linewidth]{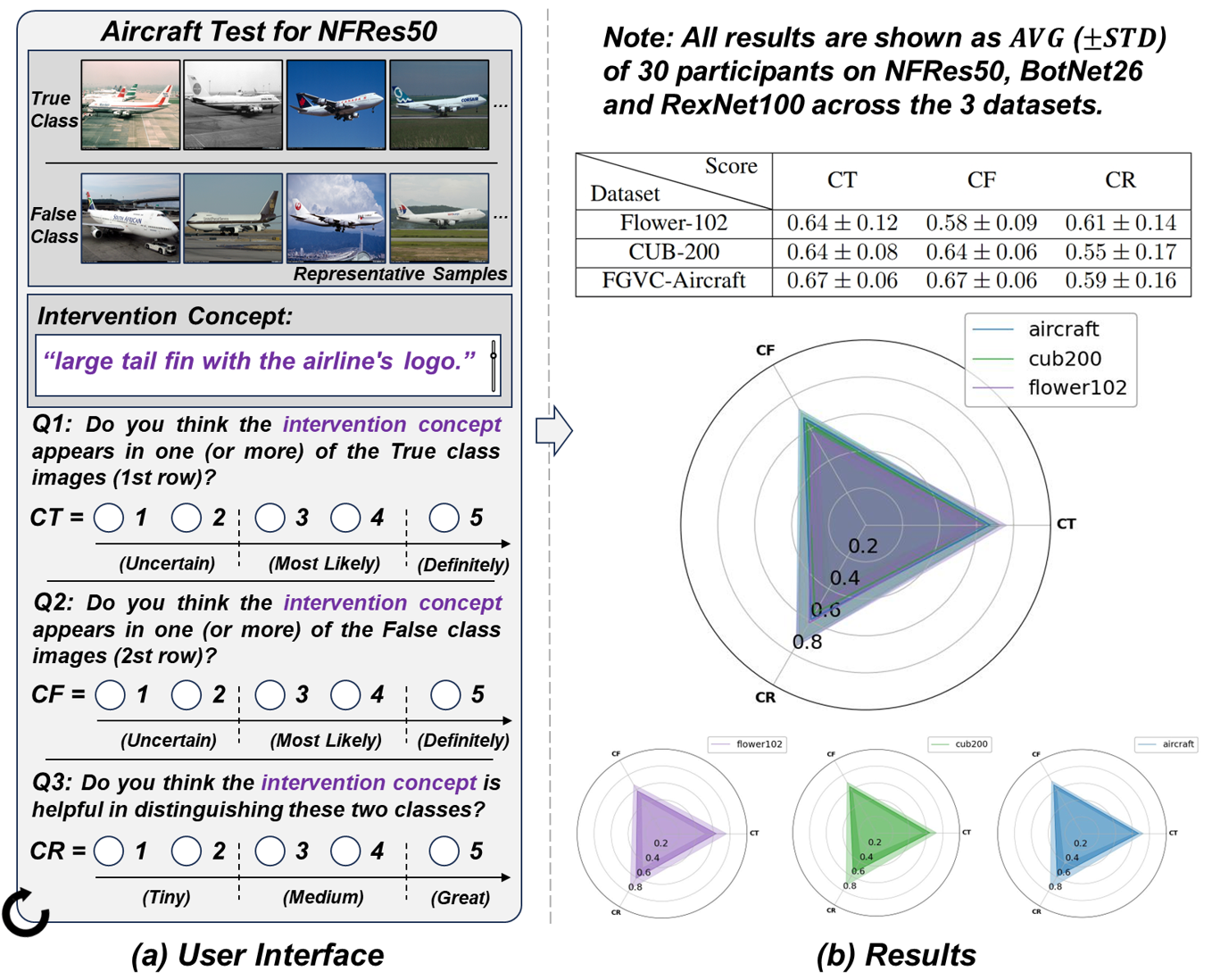}
		\caption{\small{The interface was polished, but prompts and evaluation criteria remained unchanged. Scores were normalized, with values above $0.5$ indicating consistency with the evaluation rule. The results show that our intervention concepts align with human perception and aid in classifying confused classes.}}
		\label{fig:five}
		\vspace{-0.6 cm}
	\end{figure}

	\noindent\textbf{User Study.}  
	We tested whether the concepts suppressed by CBM-HNMU are the same cues that humans find misleading.  
	The study spanned the \textit{Flower-102}, \textit{CUB-200}, and \textit{FGVC-Aircraft} test sets.  
	For each pair of confused class we clustered its images into ten groups and retained one exemplar per cluster.  
	Thirty paid, non-expert volunteers examined each exemplar along with its intervention concept and scored the concept as (i) true-class cue (CT), (ii) false-class cue (CF), or (iii) a semantically related cue shared by both classes (CR). As shown in Fig.\,\ref{fig:five}, the mean confidence for all three score exceeds~$0.5$, and the CT and CF scores are nearly identical. This parity, together with the high CR confidence, indicates that CBM-HNMU targets exactly the visual hints that mislead both the network and human observers, validating the choice of intervention concepts.
	
	%The results show that the user's confidence in the correlation (CT and CF) between the natural language concepts selected for intervention by CBM-HNMU and the visual concepts in the confusing class samples and the possible classification errors between confusing classes (CR) are generally higher than $0.5$, and the matching confidence for the true class and the false class is basically consistent. This further shows that the intervention concepts selected by the proposed method are generally the misleading concepts in the confusing classes and are consistent with human visual perception.
	
	\section{Conclusion}
	We propose a novel human neural network mutual understanding framework that combines visual concepts and natural language concept bottleneck. Based on CBM-HNMU, we can intervene in the black box based on understanding the black box prediction basis to improve the generalization performance of the black box while maintaining interpretability. By combining LLMs with pre-trained VLM and unsupervised concept extraction method, CBM-HNMU can complete automated black-box intervention beyond the sample level without additional labor costs and provide enriched intervention explanations. 
	
	\section*{Acknowledgments}
	This work was supported by grants from the Natural Science Foundation of Shanxi Province (2024JCJCQN-66), Science and Technology Commission of Shanghai Municipality (NO.24511106900), Key R\&D Program of Zhejiang (2024SSYS0091) and is partially supported by the National Natural Science Foundation of China under grant Nos. 62072358 and 62072352.
	
	%\newpage
	{\small
		\bibliographystyle{ieeenat_fullname}
		\nocite{*}
		\bibliography{references.bib}
	}
	
	\appendix
	\renewcommand\thesection{\Alph{section}}
\renewcommand*{\thefigure}{\Roman{figure}}
\renewcommand*{\thetable}{\Roman{table}}
\setcounter{figure}{0}
\setcounter{table}{0}

%\newpage
\section*{Appendix}

\noindent \textbf{Outline}

\noindent A. Concept Replacement 

\noindent B. More Evaluation on Approximation and Intervention

\noindent C. Intervention Concept Visual Masking 

\noindent D. Hardware and Software Settings 

\noindent E. Parameter Settings 

\noindent F. Discussion

\noindent G. Limitations 

\noindent H. Visualizations and Explanations \\

The organization of supplementary material is as follows: Section \ref{sec:2} introduces the concept replacement method based on additional concept search sets. Section \ref{sec:3} illustrates the optimization of predictions before and after intervention within an interpretable CBM, along with an analysis of its approximation to a black-box model for generalization. Also, the change in black-box classification accuracy for non-intervened classes before and after the intervention is reported.
%We also compare the black-box classification accuracy of the non-intervention class before and after the intervention. 
Section \ref{sec:4} details an intervention-based concept masking experiment for vision-related tasks. Sections \ref{sec:5} through \ref{sec:8} provide further experimental details and discussions. Finally, Section \ref{sec:1} presents additional visualizations of intervention explanations. To address the concept bottleneck limitations.

%\textit{\textbf{The attached file provides the intervention classes of three models on three fine-grained classification datasets and the corresponding selected intervention concepts.}}

\section{Concept Replacement}\label{sec:2}

It is also noteworthy that simply deleting concepts may not resolve classification errors caused by conceptual bottleneck limitations. To address this, we propose a concept replacement method that leverages a search set of additional concepts.

\begin{figure*}[htp]
	\centering
	\includegraphics[width=0.9\linewidth]{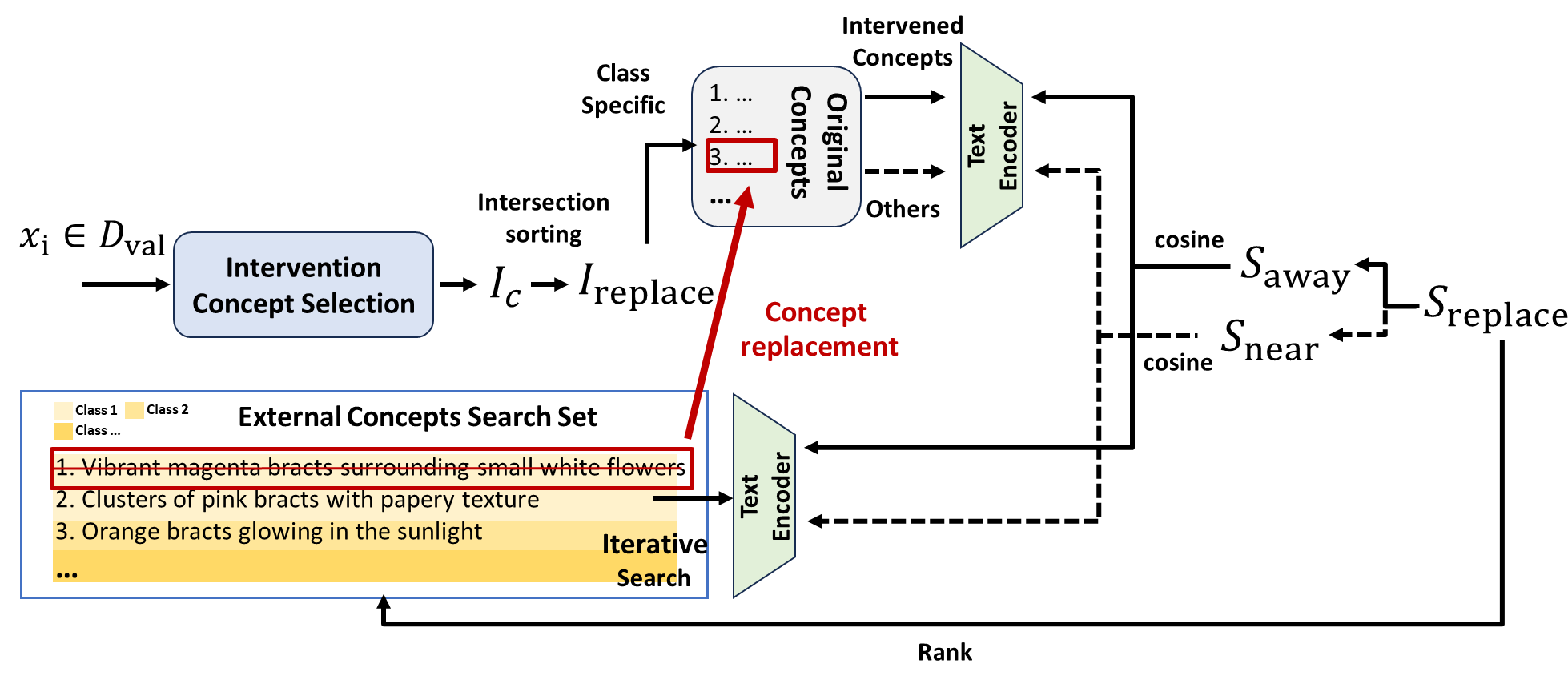}
	\caption{Concepts replacement framework. \textbf{Intervention Concept Selection:} Obtain the concepts that require intervention for each confused class. \textbf{Intersection Sorting:} The intervening concepts for all confused classes are intersected, and the concepts that need to be replaced are obtained by sorting the number of occurrences. \textbf{Concept Replacement:} Iteratively search for candidates in the External Concepts Search Set that are far away from the replacement concept but similar to other positive concepts of the same class, sort according to the $S_{\mathbf{replace}} = S_{\mathbf{near}} - S_{\mathbf{away}}$, and select the optimal concept replacement.}
	\label{fig:III}
	\vspace{-0.6 cm}
\end{figure*}

The framework of the replacement method is illustrated in Figure \ref{fig:III}. It begins by identifying concepts that require intervention within each confusion class through concept intervention. These concepts are then ranked based on their frequency of occurrence, and the
\text{top $\bar{q}$} most frequently occurring concepts are selected for replacement. For each identified concept, positive concepts that do not require modification are also determined. Simultaneously, a replacement concept is selected from an external search set, with the objective of positioning it as far as possible from the embedding of negative concepts while bringing it closer to positive concepts. Cosine similarity is utilized to measure expression similarity, allowing us to score potential replacement concepts effectively.

\begin{figure}[htp]
	\centering
	\includegraphics[width=\linewidth]{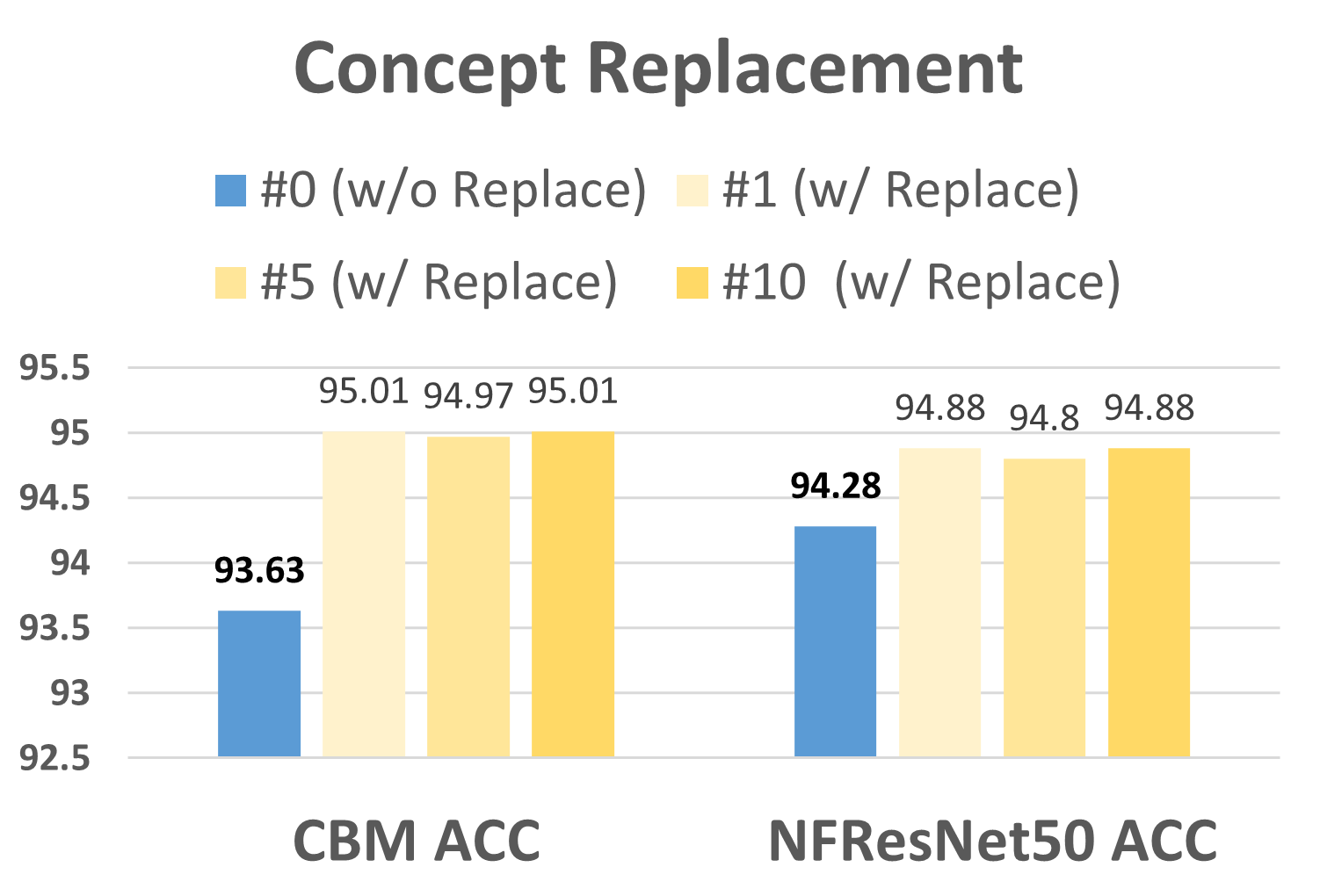}
	\caption{Comparison of accuracy before and after the concept replacement of NFResNet50 and the corresponding CBM on Flower-102.}
	\label{fig:IV}
	\vspace{-0.6 cm}
\end{figure}

Figure \ref{fig:IV} presents the accuracy comparison of NFResNet50 on Flower-102 before and after concept replacement. The results demonstrate that replacing different numbers of concepts leads to improved test accuracy for both the black-box baseline and the corresponding CBM.

\renewcommand{\arraystretch}{1.5} 
\begin{table*}[htp]
	\centering
	\resizebox{0.75\linewidth}{!}{
		\begin{tabular}{ccccccc}
			\hline
			\multirow{2}{*}{Models} & \multicolumn{2}{c}{Flower-102} & \multicolumn{2}{c}{CUB-200} & \multicolumn{2}{c}{FGVC-Aircraft} \\
			\cline{2-7}
			& $w/o$ $INT$ & $w/$ $INT$  & $w/o$ $INT$ & $w/$ $INT$  & $w/o$ $INT$ & $w/$ $INT$ \\
			\hline
			NFResNet50 & 93.63 & 93.75 ($\uparrow 0.12$)  & 61.90 & 62.70 ($\uparrow 0.80$) & 65.80 & 66.67 ($\uparrow 0.87$) \\
			Vit\_Small & 80.10 & 80.28 ($\uparrow 0.18$) & 54.10 & 54.15 ($\uparrow 0.05$) & 62.95 & 63.55 ($\uparrow 0.60$) \\
			ResNeXt26 & 89.16 & 90.69 ($\uparrow 1.53$) & 60.15 & 60.50 ($\uparrow 0.35$) & 63.79 & 64.69 ($\uparrow 0.90$) \\
			GCVit\_Base & 92.84 & 93.26 ($\uparrow 0.42$) & 77.10 & 77.15 ($\uparrow 0.05$) & 68.98 & 69.91 ($\uparrow 0.93$) \\
			\hline
		\end{tabular}
	}
	\caption{Comparison of classification accuracy with $P_{CBM}$ ($w/o$ $INT$) and $\tilde{P}_{CBM}$ ($w/ INT$).}
	\label{tab:I}
	\vspace{-0.3 cm}
\end{table*}

\begin{figure*}[htp]
	% \vspace{-0.6 cm}
	\centering
	\includegraphics[width=1.0\textwidth]{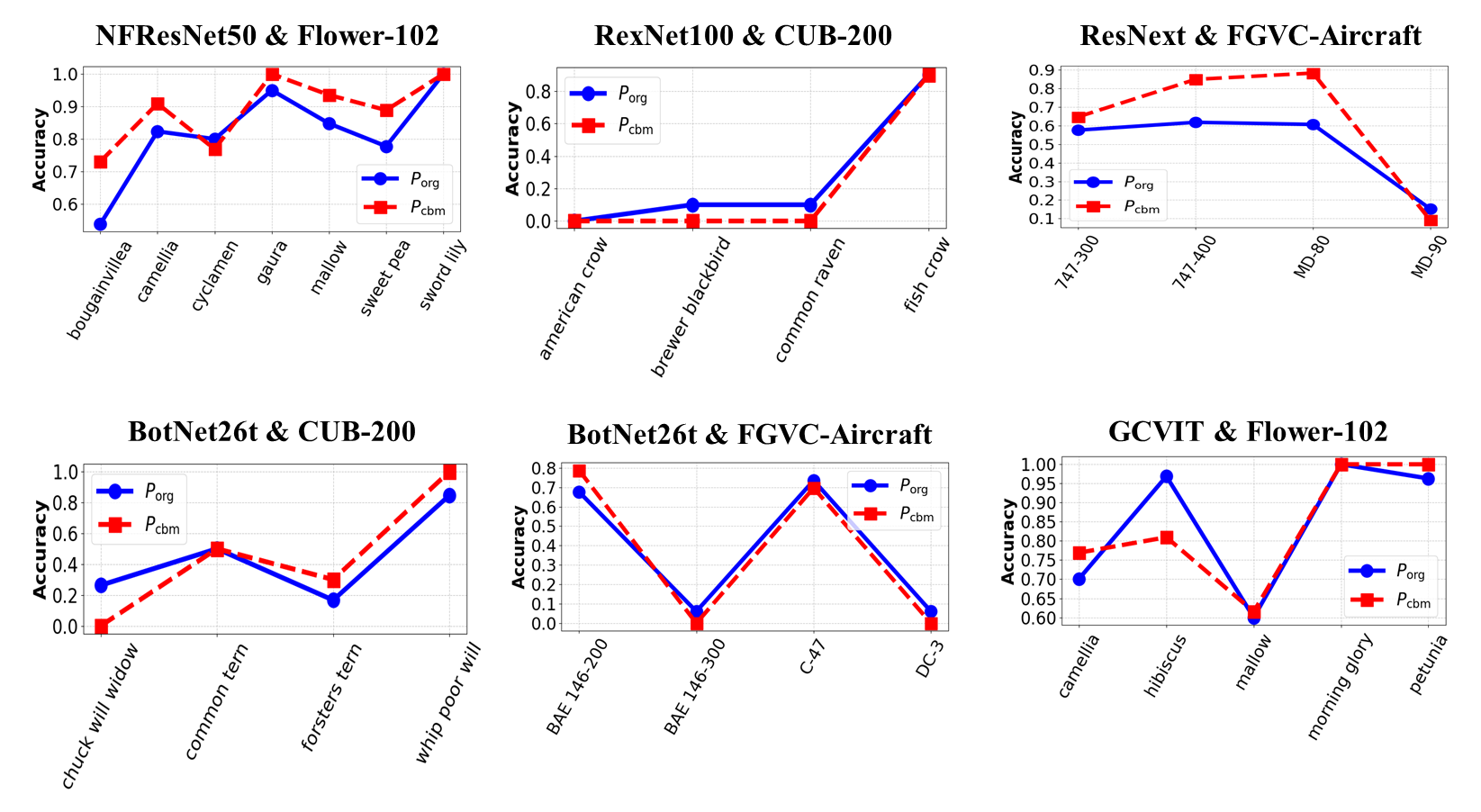}
	\caption{Confusion class accuracy comparison curve between part of the CBM approximate reasoning structure and the original black box. The y-axis is the prediction accuracy of the class, and the x-axis is the corresponding intervention class ($\Gamma$).}
	\label{fig:II}
	\vspace{-0.6 cm}
\end{figure*}

\begin{figure*}[htp]
	%\vspace{-0.15 cm}
	\centering
	\includegraphics[scale=0.9]{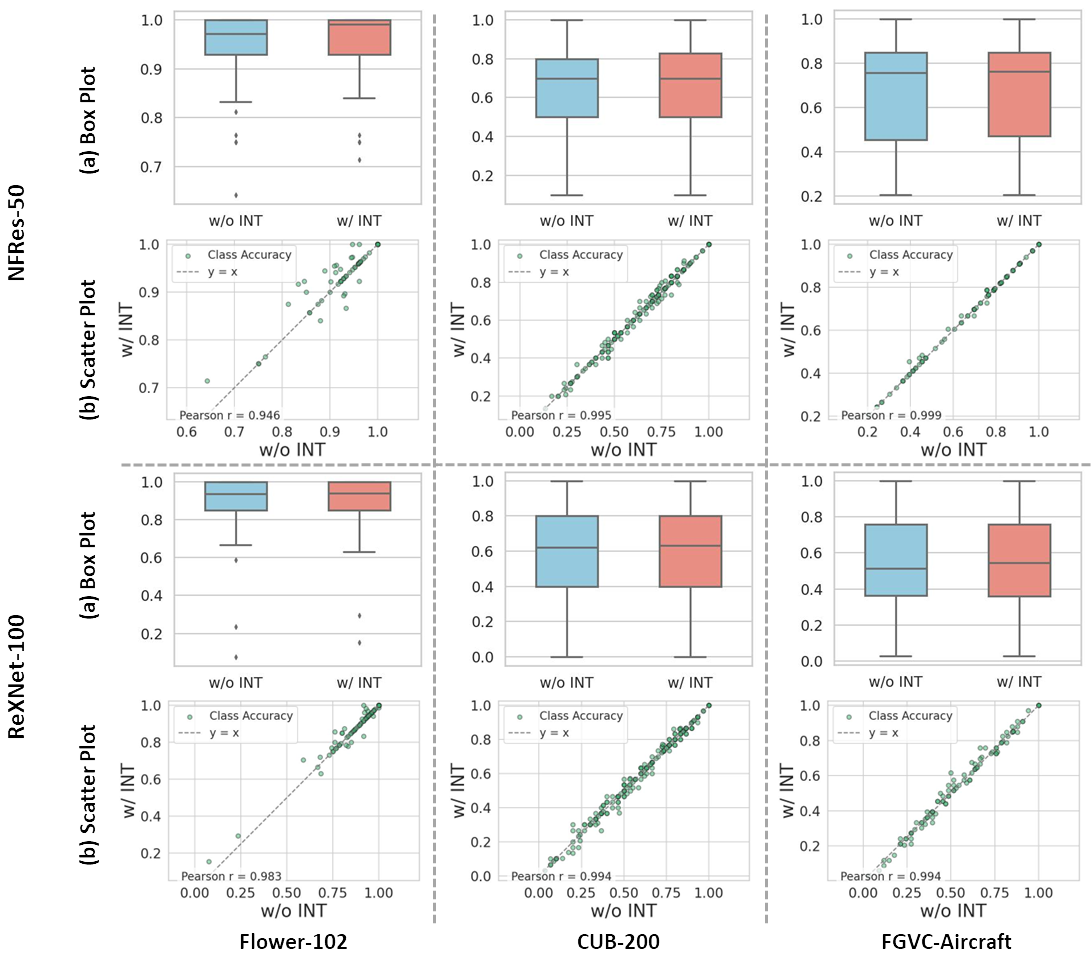}
	\caption{ \small{Before and after intervention in non-intervention classes.}}
	\label{fig:bs}
	\vspace{-0.6 cm}
\end{figure*}

\section{More Evaluation on Approximation and Intervention}\label{sec:3} 
We also conduct a class-based accuracy analysis on the approximate black-box CBM inference structure and compare it with the original black-box model. Figure \ref{fig:II} presents a subset of accuracy comparison curves for confusing classes between the CBM-based approximate reasoning structure and the original black box. Additionally, Table \ref{tab:I} reports improvements achieved by CBM inference after applying a concept weight matrix in CBM-HNMU.
The results indicate that the classification accuracy curve of CBM on intervention classes generally aligns well with that of the original black box. This suggests that CBM can effectively approximate the reasoning logic of the black box within locally confused classes. Furthermore, we evaluate the accuracy of the concept intervention algorithm based on CBM. The findings reveal that nearly all interventions help correct concept-related reasoning errors, leading to improvements in both class-level and global accuracy.

% \red In Section 3.6, we use $\mathbf{pr}$ in Eq. (9) to transfer intervention knowledge while preserving the distribution of non-target classes. As a result, their accuracy remains largely unaffected. We report the accuracy change on non-intervention classes for two models across three datasets in Figure \ref{fig:bs}. The results show stable performance closely correlated with pre-intervention accuracy (scatter plot), along with a modest overall improvement (box plot).\black
To preserve the predicted distribution over non-target classes when transferring intervention knowledge, we apply the operator -- $\mathbf{pr}$, to reassign only the residual probability mass from non-target classes to the intervention class, as defined in Eq. (9). Figure \ref{fig:bs} reports the accuracy shift for non-target classes across both models and all three datasets. The results show that performance is stable and closely aligned with pre-intervention accuracy (scatter plot), with a slight overall improvement (box plot).
% On the other hand, to preserve the predicted distribution of non-target classes when transferring the intervention knowledge, we use $\mathbf{pr}$ to only assign the remaining probability values of non-target classes to the intervention class using Eq. (9) in Section 3.6 of the paper. Therefore, the accuracy of non-target classes is basically unaffected after the intervention. We report the change in accuracy of non-target classes for the two models on the three datasets in Figure \ref{fig:bs}. The results show that the performance is stable and closely related to the accuracy before the intervention (scatter plot), while slightly improving overall (box plot).
% \red In Section 3.6, we use $\mathbf{pr}$ in Eq. (9) to transfer intervention knowledge while preserving the distribution of non-target classes. As a result, their accuracy remains largely unaffected. As suggested, we report the accuracy change on non-intervention classes for two models across three datasets in Figure \ref{fig:bs}. The results show stable performance closely correlated with pre-intervention accuracy (scatter plot), along with a modest overall improvement (box plot).\black

\begin{figure}[htp]
	\vspace{-0.6 cm}
	\centering
	\includegraphics[width=1\linewidth]{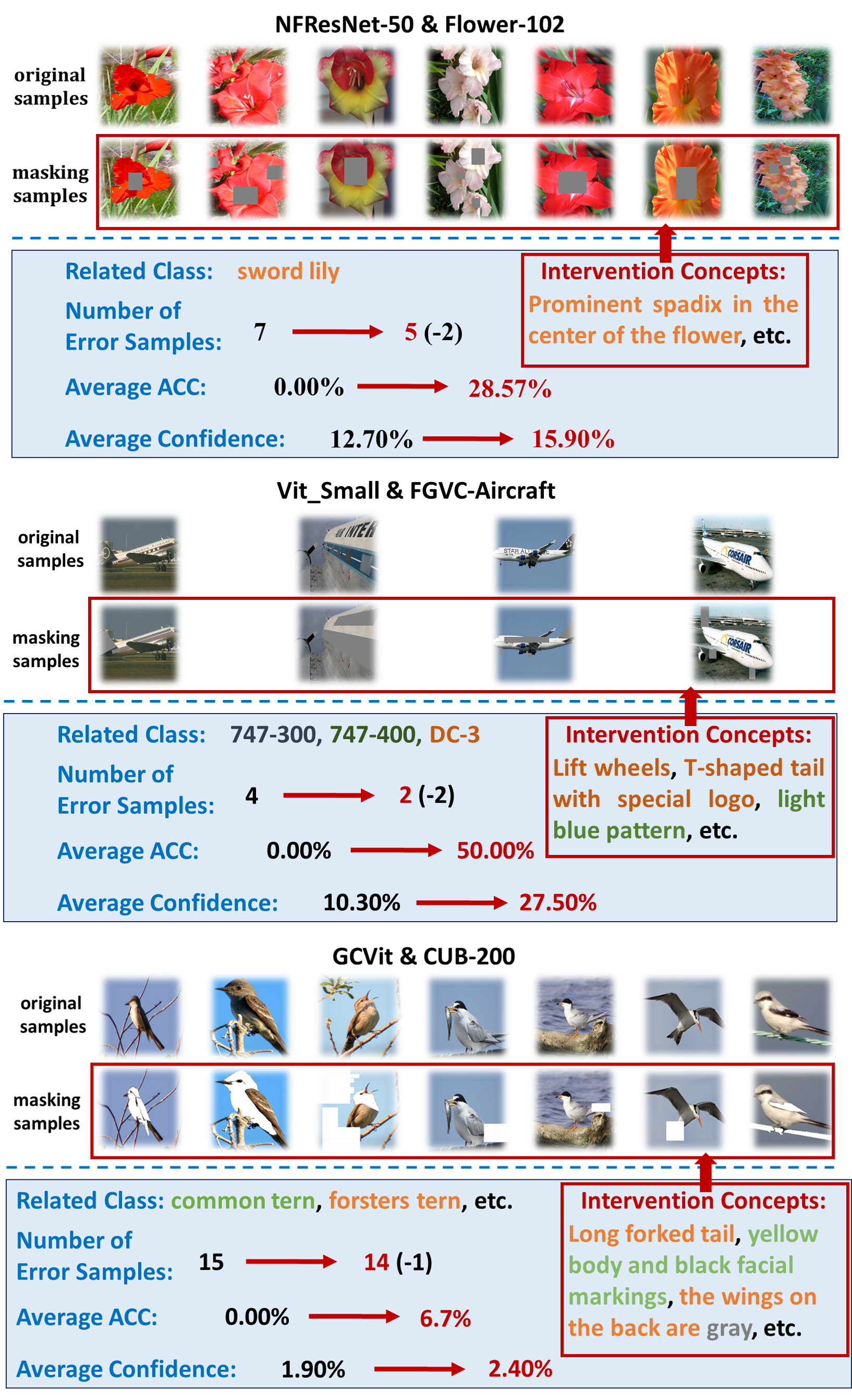}
	\caption{Visual related concepts masking. Intervention concepts and class name having the same color represent that the intervention concept belongs to the corresponding class, and the visual mask of samples belonging to this class is labeled according to these concepts.}
	\label{fig:V}
	\vspace{-0.6 cm}
\end{figure}
%\vspace{-0.3 cm}

\begin{figure*}[htp]
	\centering
	\includegraphics[scale=0.50]{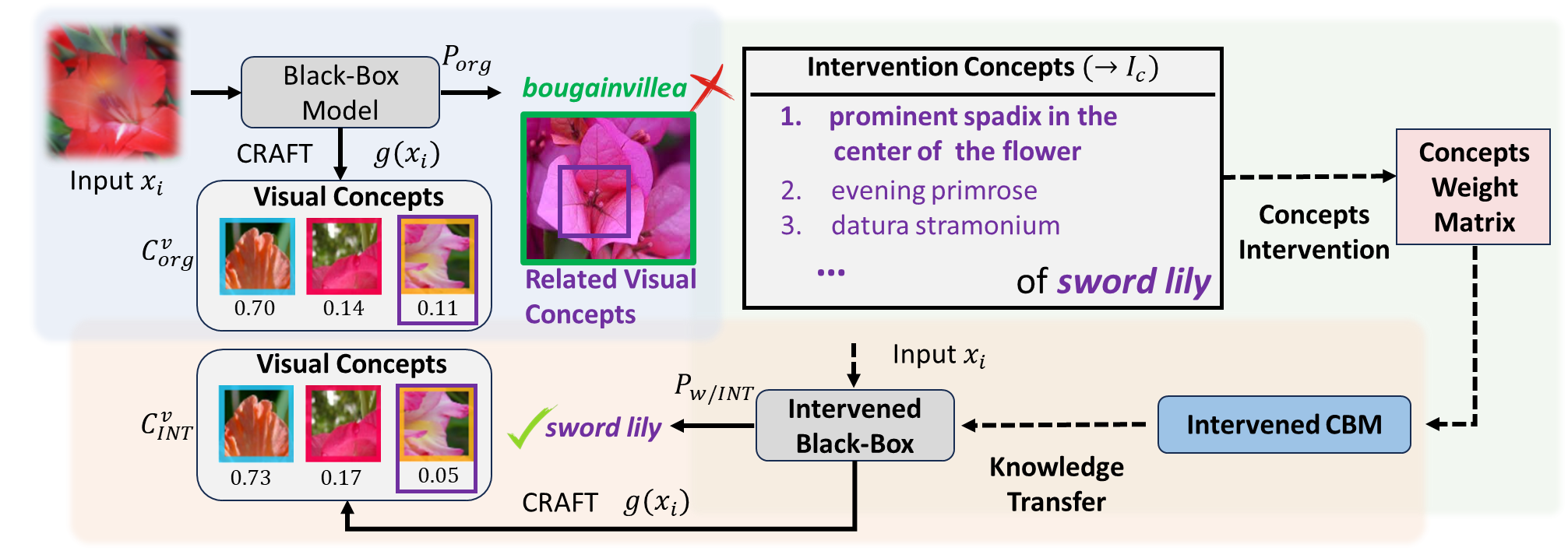}
	\caption{More visualization of NFResNet50 on Flower-102.}
	\label{fig:1}
	\vspace{-0.3 cm}
\end{figure*}

\section{Intervention Concept Visual Masking}\label{sec:4}

We use OpenAI-CLIP as the concept communication module, enabling CBM-HNMU to align natural language intervention semantics with corresponding visual concept changes before and after intervention. To validate this alignment, we conduct visual masking experiments on intervention-related concepts across different models and datasets.

First, we identify a specific class of intervention concepts in the black-box model and extract the top-ranked natural language concepts with the highest intervention scores ($S_{\text{nT}}/S_{\text{pF}}$)  that are also visually relevant. Next, we collect all samples from that class that the black-box model misclassifies. Finally, we apply pixel-based masking to mask the corresponding semantic information associated with the intervening natural language concept, generating new sample inputs.

These modified samples are then fed into the original black-box model to obtain new predictions and confidence scores, and compare with the original results. As shown in Figure \ref{fig:V}, in the first example, we visually masked $7$ misclassifies samples of \textit{sword lily} on Flower-102 using NFResNet-50. Based on the vision-related concepts identified in the intervention’s natural language description, we masked the central pixels of the flowers in these samples.

In the second example, we performed a similar procedure on four misclassified samples spanning three aircraft classes (\textit{747-300, 747-400 and DC-3}). Notably, when conducting visual masking experiments on CUB-200 using GCVit, the intervention concept exhibited the semantic feature of gray. To account for this, we applied a pure white mask to cover the pixels corresponding to the intervention concept.

\section{Hardware and Software Settings}\label{sec:5}

All experiments in this work are conducted on Ubuntu 20.04 within an Anaconda3 virtual environment, using NVIDIA 3090 (24GB) GPU. This setup allows us to provide integrated environment resources, including code and datasets, in a public remote hub in the future. 

The pre-trained network weights and datasets referenced in the paper are publicly available resources. We will include download links and deployment instructions in subsequent packaged code. Additionally, for the various methods cited in the paper, we will provide links along with detailed deployment guidelines.

\section{Parameter Settings}\label{sec:6}

All baseline models ($P_{org}$ $w/o$ $INT$) use weights pre-trained on ImageNet-1K, and are tuned on the corresponding experimental datasets with $50$ epoch. During the fine-tuning process, the learning rate is set to $1e^-4$. Both confusing classes selection, local approximation are performed on the corresponding $D_{val}$. During local approximation, learning rate is set to $1e^-4$ and epoch is set to $200$. Concepts intervention is still performed on the $D_{val}$ and only execute the \textcolor{blue}{Algorithm 1} in the paper once to modify the concept weight matrix ($W$) of the corresponding $P_{CBM}$. 

Knowledge transfer requires setting different distillation temperatures for teachers and student, where the teacher model $P_t$ is spliced by the frozen original black box ($P_t^1$) and locally approximated CBM ($P_t^2$). The student model $P_s$ is the original black box. The distillation temperature $T_1$ of $P_t^2$ is $2.0$, and the distillation temperature $T_2$ of $P_s$ is $1.5$. Knowledge transfer is performed on the $D_{val}$ for $10$ epoch with learning rate $3e^-7$. The maximum number of intervention concepts varies depending on the datasets. It is recommended to set it between $10$ and $100$. The number can be adjusted according to the intervention effect (we take the optimal value in multiple groups of experiments). 

\section{Discussion}\label{sec:7}

In this paper, we demonstrate the effectiveness of the CBM-HNMU approach combined with gradient-based intervention. In fact, according to the human-understandable intervention concepts provided by CBM-HNMU, we can even manually select the visual part corresponding to the concept to quickly intervene and determine the harmful concept dependence of the model on the data domain. Secondly, CBM-HNMU bridges the black box and interpretable structures, integrates visual and language modalities, and provides intuitive model explanations for easy understanding. The method's explanation-based intervention effectively identifies the recognition patterns and biases inherent in black-box models, laying the foundation for building interpretable classification networks in the future.

\section{Limitations}\label{sec:8}

CBM-HNMU relies on both visual concepts and natural language concepts extracted from the black box. Although many unsupervised methods can be used to efficiently extract the concepts of the corresponding model and even give attribution explanations, and using LLMs can quickly obtain the concept bottleneck of the corresponding datasets, inevitably due to 1) limitations of concept extraction methods, such as concept extraction and explanation methods that are not model-oriented, it will lead to situations where concepts cannot be well connected to samples. 2) The hallucination phenomenon of LLMs may produce a large number of abstract concepts. Abstract concepts and concrete concepts have little impact on the expression of concept bottleneck and OpenAI-CLIP can be also used to connect abstract concepts with visual feature. However, it is disadvantageous for human to understand the model intervention process and error correction explanation based on the relationship between abstract concepts and visual concepts. Problem 1) can be solved by applying model-oriented concept extraction methods, such as the \textit{Model-Oriented Concept Extraction} (MOCE). For question 2), manual verification is a more compromised method, which can save most of the time while ensuring the quality of the concept.

\begin{figure*}[htp]
	\centering
	\includegraphics[scale=0.50]{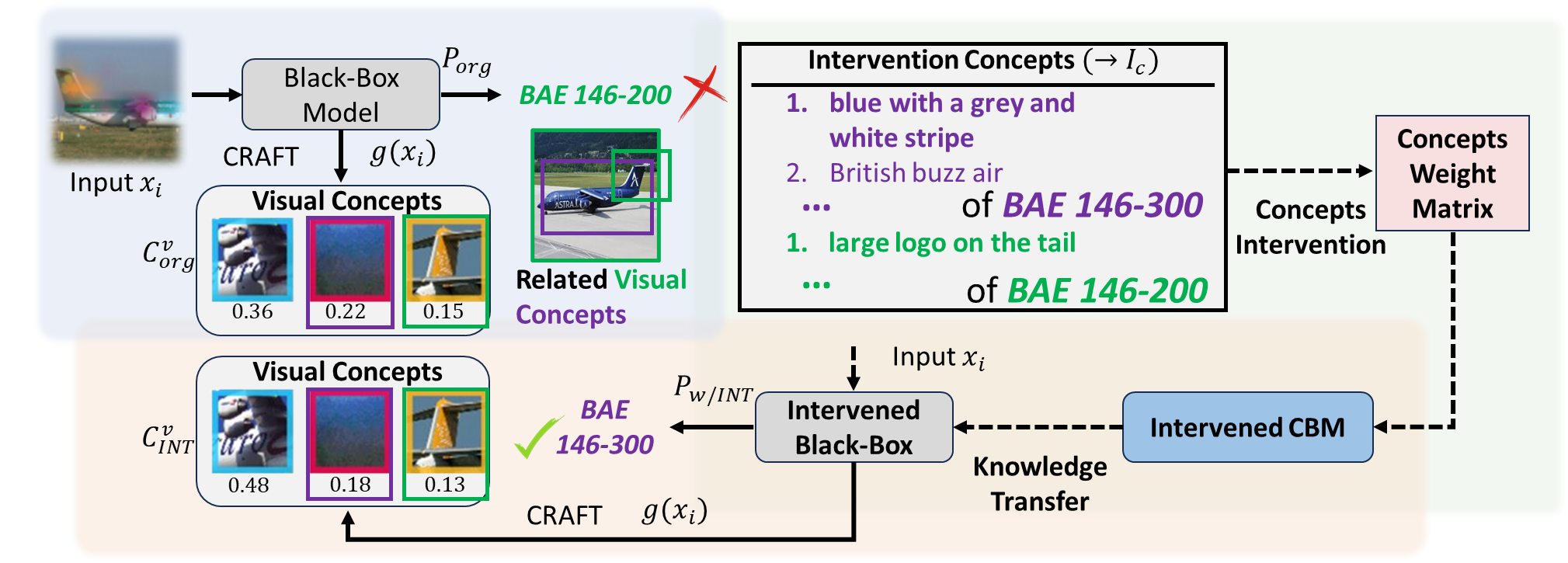}
	\caption{More visualization of NFResNet50 on FGVC-Aircraft.}
	\label{fig:2}
	\vspace{-0.3 cm}
\end{figure*}

\begin{figure*}[hbp]
	\centering
	\includegraphics[scale=0.50]{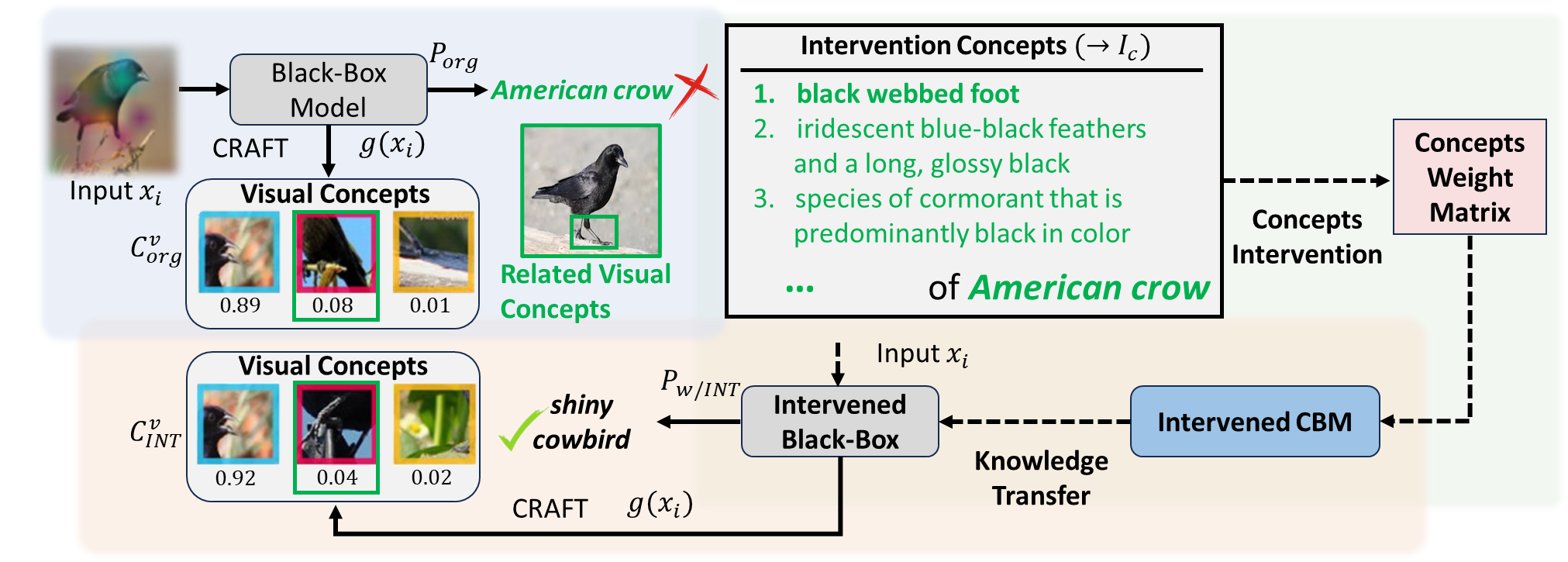}
	\caption{More visualization of NFResNet50 on CUB-200.}
	\label{fig:3}
	\vspace{-0.3 cm}
\end{figure*}

\begin{figure*}[htp]
	\centering
	\includegraphics[scale=0.50]{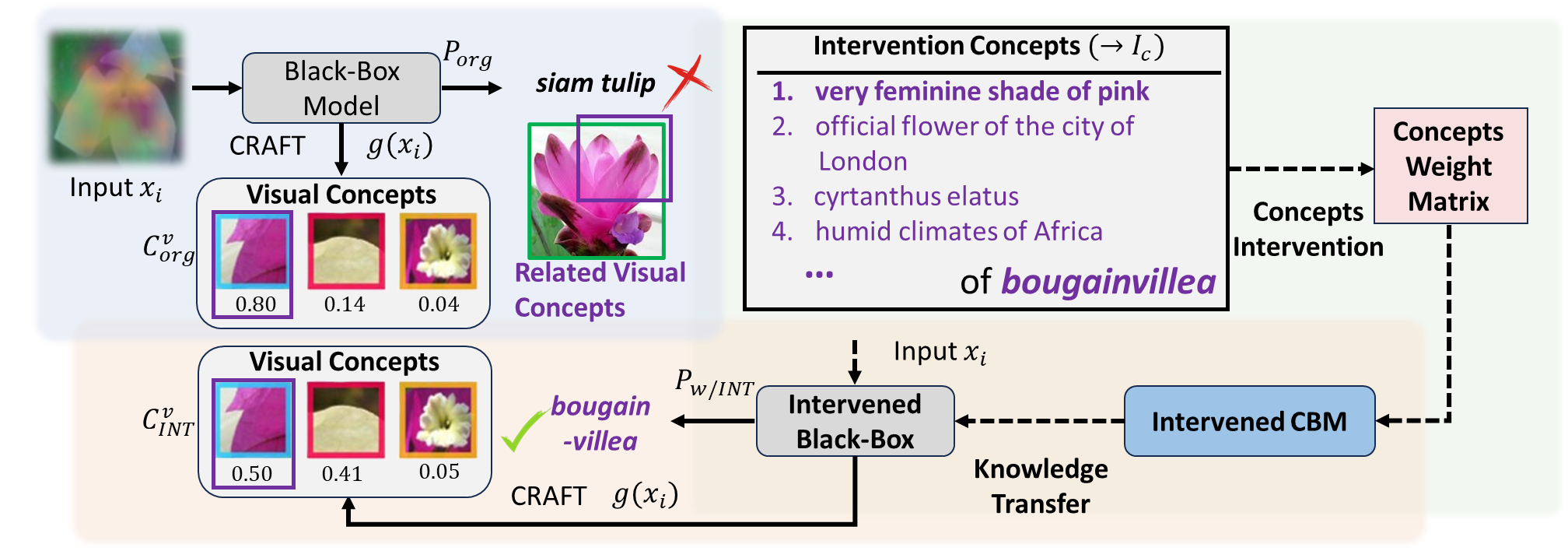}
	\caption{More visualization of BotNet26 on Flower-102.}
	\label{fig:4}
	\vspace{-0.3 cm}
\end{figure*}

\section{Visualization and Explanation}\label{sec:1}

\begin{figure*}[hbp]
	\centering
	\includegraphics[scale=0.50]{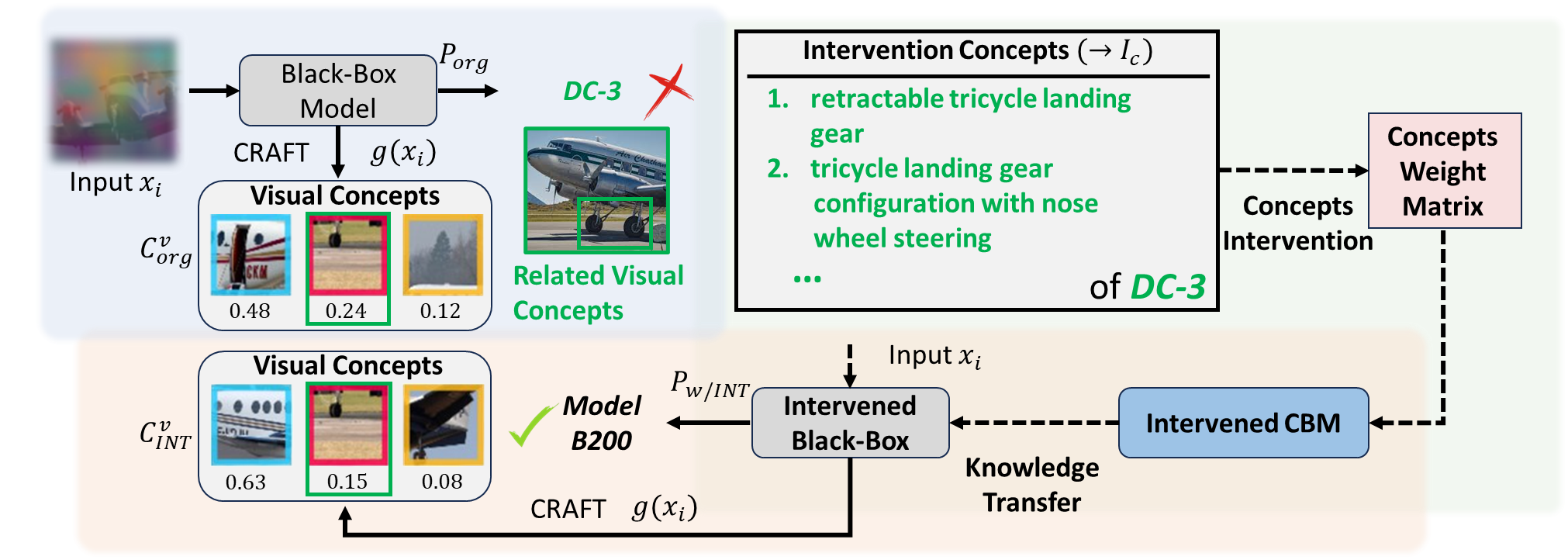}
	\caption{More visualization of BotNet26 on FGVC-Aircraft.}
	\label{fig:5}
	\vspace{-0.3 cm}
\end{figure*}

\begin{figure*}[htp]
	\centering
	\includegraphics[scale=0.50]{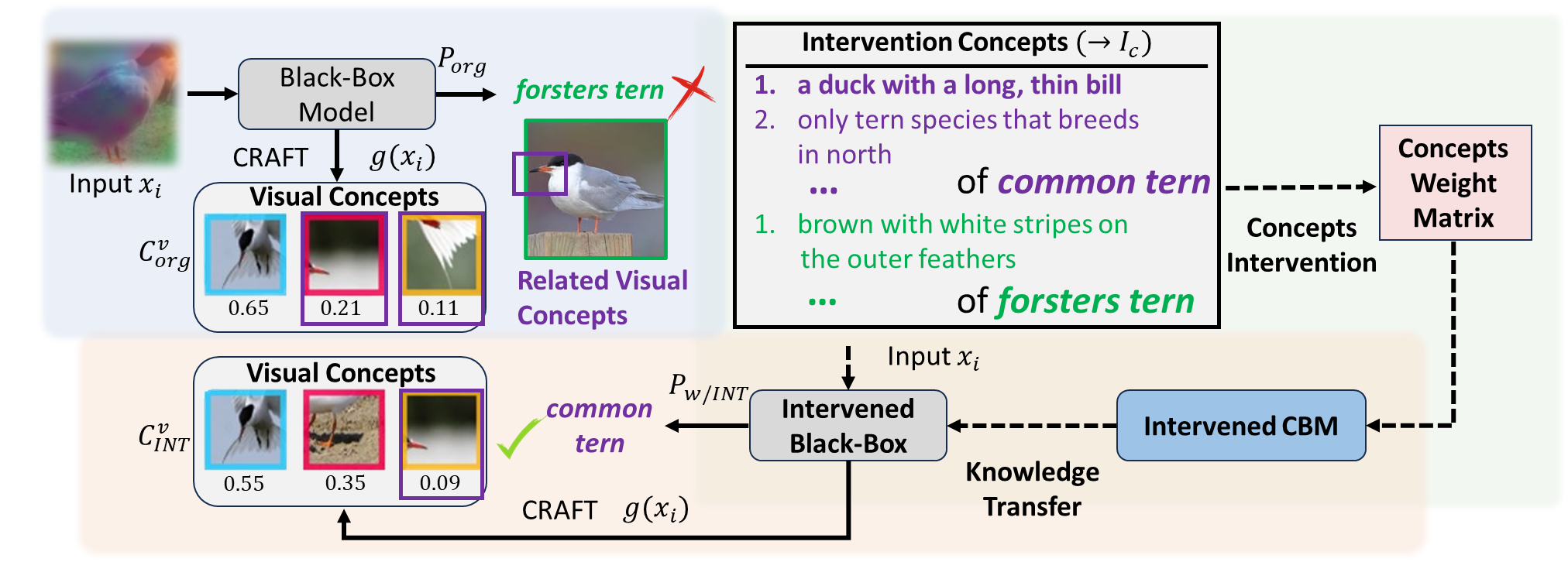}
	\caption{More visualization of BotNet26 on CUB-200.}
	\label{fig:6}
	\vspace{-0.3 cm}
\end{figure*}

\begin{figure*}[hbp]
	\centering
	\includegraphics[scale=0.50]{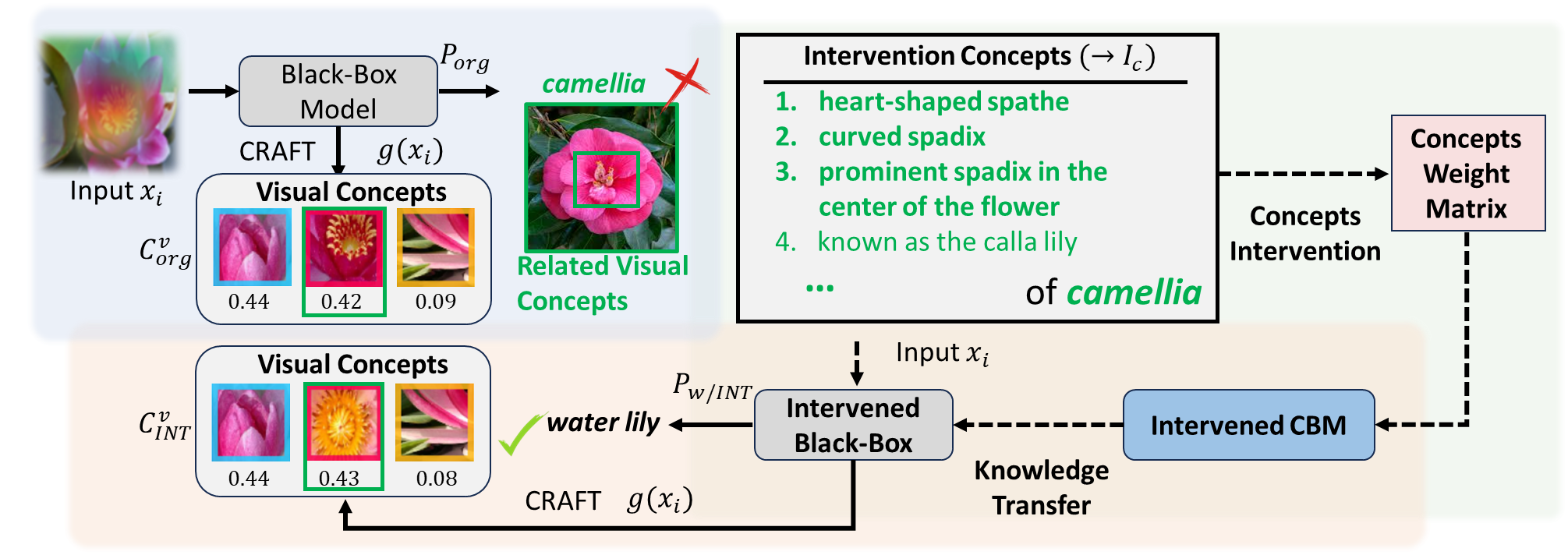}
	\caption{More visualization of RexNet100 on Flower-102.}
	\label{fig:7}
	\vspace{-0.3 cm}
\end{figure*}

\begin{figure*}[htp]
	\centering
	\includegraphics[scale=0.50]{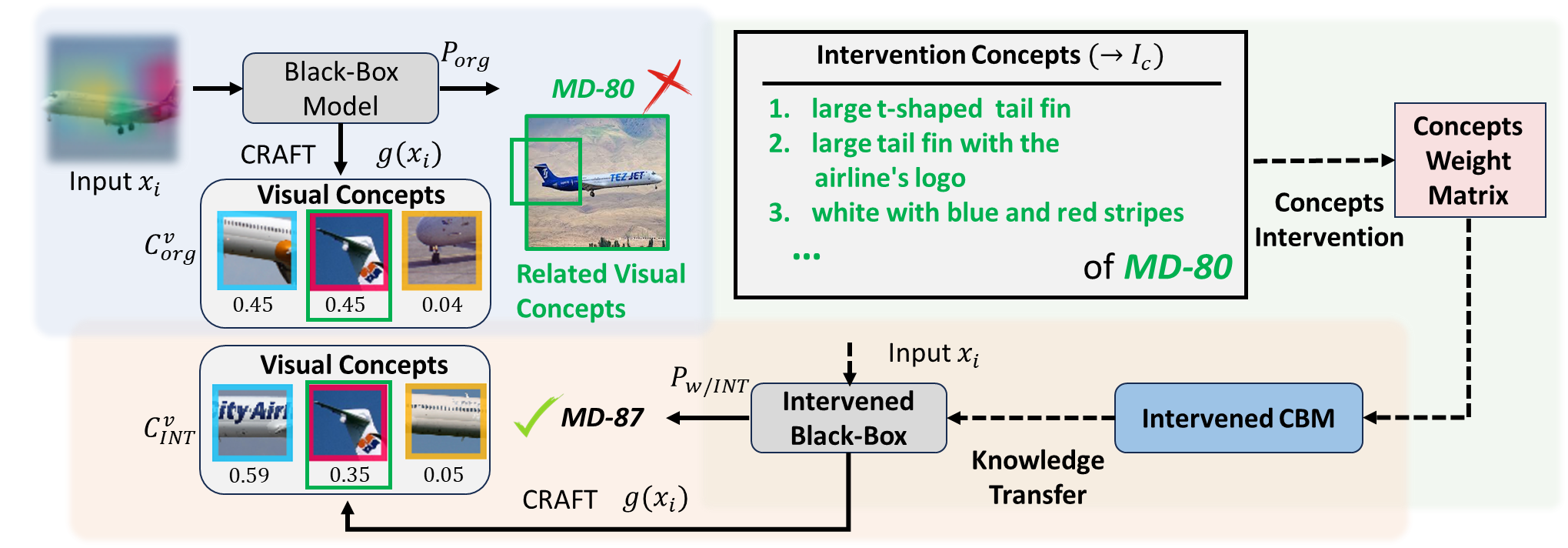}
	\caption{More visualization of RexNet100 on FGVC-Aircraft.}
	\label{fig:8}
	\vspace{-0.3 cm}
\end{figure*}

\begin{figure*}[hbp]
	\centering
	\includegraphics[scale=0.50]{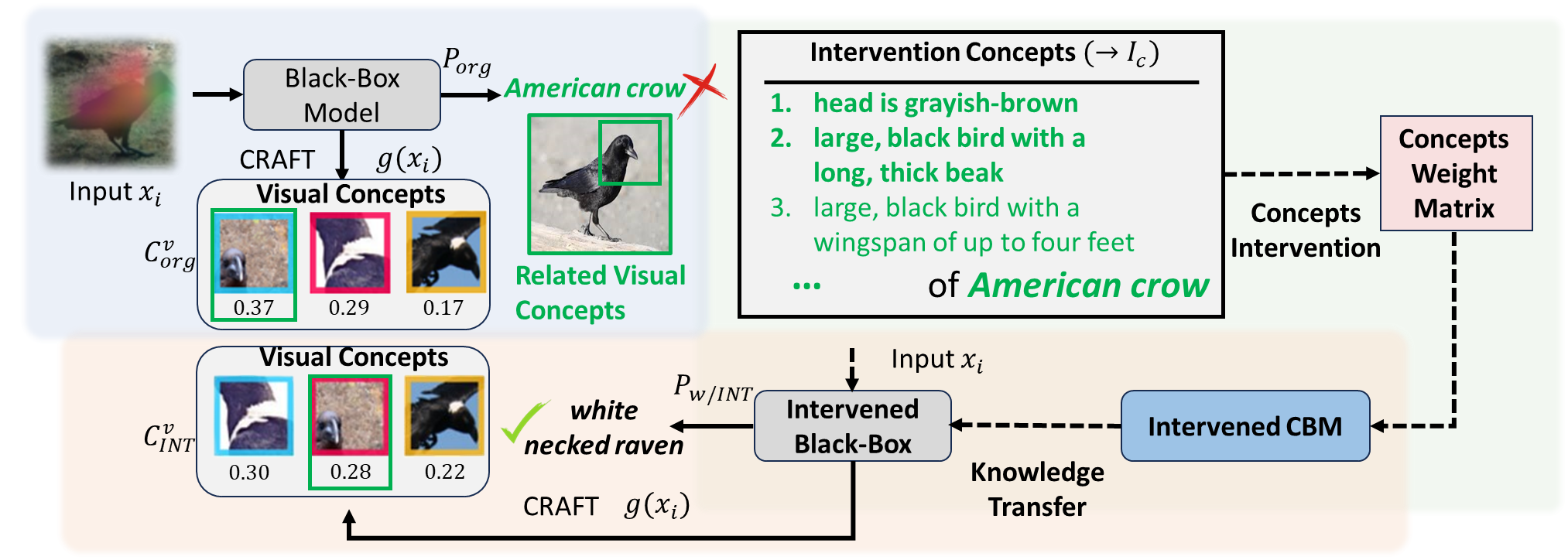}
	\caption{More visualization of RexNet100 on CUB-200.}
	\label{fig:9}
	\vspace{-0.3 cm}
\end{figure*}

We begin with additional visualizations of intervention-based explanations, as shown in Figures \ref{fig:1} -- \ref{fig:9}. These illustrations highlight the relationship between natural language intervention concepts and changes in black-box visual representations before and after intervention. We present results using CBM-NHMU on NFResNet50, BotNet26, and RexNet100, with interventions applied to Flower-102, CUB-200, and FGVC-Aircraft. Notably, the samples are randomly selected from a subset where the black-box model’s original classification errors on the test set are corrected after intervention. Each corrected sample includes at least one pre- or post-correction class associated with the confused classes, ensuring a clear visual link to the intervention concept (see ``Coverage" in the manuscript).

Before detailing each intervention explanation example, we first clarify the key components in each visualization. In each example, the upper-left image represents the CRAFT concept attribution of the input image to the post-intervention black-box model. Recall $P_{org}$ and $P_{w/INT}$ denote the classification predictions of the black-box model before and after the intervention, respectively.

Incorrect class predictions (i.e., those made by the original black-box model) are marked \textbf{\textcolor{dgreen}{green}} if they belong to the confused classes ($\Gamma$) and \textbf{\textcolor{black}{black}} otherwise. The correct class prediction (i.e., the black-box model’s output after intervention) is marked \textbf{\textcolor{purple}{purple}} if it falls within the confusion category; otherwise, it is also marked by \textbf{\textcolor{black}{black}}. In each visualization below, real example images corresponding to incorrectly predicted classes are displayed. These examples help illustrate why the black-box model misclassified the input and how the intervention concept corrects the error. By examining images from confused categories alongside intervention concepts, users can better understand the reasoning behind the intervention. Intervention concepts and their semantically related visual counterparts in the confused classes are highlighted with \textbf{\textcolor{purple}{purple outlines}}, whereas relevant concepts from the incorrectly predicted class are enclosed in \textbf{\textcolor{dgreen}{green borders}}. We are now in the position to detail intervention explanation examples.

In Figure \ref{fig:1}, the baseline of NFResNet50 misclassifies \textcolor{purple}{\textit{sword lily}} as \textcolor{dgreen}{\textit{bougainvillea}}. The visually related intervention concept: \textcolor{purple}{\textit{"Prominent spadix in the center of the flower."}} prompts us to delete the related concept about the prominence in the center of flower to correct this error. We can find that the third most important visual concept extracted by the original black box before intervention corresponds exactly to this description, and the importance score of this concept is $0.11$. After the intervention, the black box gives a similar conceptual explanation, but the difference is that the importance score of the intervention concept dropped significantly ($\downarrow 0.06 \rightarrow 0.05$). Combining the misclassifies sample image, we can also see that the visual features corresponding to the intervention concepts are indeed easy to confuse the two classes.

In Figure \ref{fig:2}, NFResNet50 misclassifies \textcolor{purple}{\textit{BAE 146-300}} as \textcolor{dgreen}{\textit{BAE 146-200}}. The visually related intervention concept: \textcolor{purple}{\textit{"Blue with a grey and white stripe."}} and \textcolor{dgreen}{\textit{"Large logo on the tail."}} prompts us to delete the related concept about the tail and the color-related feature of aircraft to correct this error. We can find that the second and the third most important visual concept extracted by the original black box before intervention corresponds exactly to the descriptions. After the intervention, we can find that the network relies less on the corresponding color-related feature and tail of the aircraft to classify real class. The concept importance scores of color-related features and aircraft tail dropped from $0.22$ to $0.18$ and $0.15$ to $0.13$, respectively.

In Figure \ref{fig:3}, NFResNet50 misclassifies \textit{shiny cowbird} as \textcolor{dgreen}{\textit{American crow}}. The visually related intervention concept: \textcolor{purple}{\textit{"Black webbed foot."}} prompts us to delete the related concept about the foot of bird to correct this error. We can find that the second and the third most important visual concept extracted by the original black box before intervention corresponds exactly to this description. After the intervention, we can find that the network relies less on bird footsteps to classify real class.

In Figure \ref{fig:4}, BotNet26 misclassifies \textcolor{purple}{\textit{bougainvillea}} as \textit{siam tulip}. The visually related intervention concept: \textcolor{purple}{\textit{"Very feminine shade of pink."}} prompts us to delete the related concept about the pink color of flower to correct this error. We can find that the first most important visual concept extracted by the original black box before intervention corresponds exactly to this description, and the importance score of this concept is $0.80$. After the intervention, the black box gives a similar conceptual explanation, but the difference is that the importance score of the intervention concept dropped significantly ($\downarrow 0.30 \rightarrow 0.50$).

In Figure \ref{fig:5}, BotNet26 misclassifies {\textit{Model B200} as \textcolor{dgreen}{\textit{DC-3}}. The visually related intervention concept: \textcolor{dgreen}{\textit{"Retractable tricycle landing gear." and "tricycle landing gear configuration with nose wheel steering."}} prompts us to delete the related concept about the gear of aircraft to correct this error. We can find that the second most important visual concept extracted by the original black box before intervention corresponds exactly to this description. After the intervention, we can find that the network relies less on gear of the aircraft to classify real class and the original score of the corresponding visual concept dropped from $0.24$ to $0.15$.
	
In Figure \ref{fig:6}, BotNet26 misclassifies \textcolor{purple}{\textit{common tern}} as \textcolor{dgreen}{\textit{forsters tern}}. The visually related intervention concept: \textcolor{purple}{\textit{"A duck with a long, thin bill."}} prompts us to delete the related concept about the bill of bird to correct this error. We can find that the second and the third most important visual concept extracted by the original black box before intervention includes this description. After the intervention, we can find that the network relies less on bird bill to classify real class and the original score of the second most important visual concept dropped from $0.21$ to $0.09$. The third most important visual concept even disappears.
	
In Figure \ref{fig:7}, RexNet100 misclassifies \textit{water lily} as \textcolor{dgreen}{\textit{camellia}}. The visually related intervention concept: \textcolor{dgreen}{\textit{"heart-shaped spathe.", "curved spadix." and "prominent spadix in the center of the flower."}} prompts us to delete the related concept about the pistil of flower to correct this error. We can find that the second most important visual concept extracted by the original black box before intervention includes corresponding features. However, the stamen feature given in concept $2$ can easily be confused between the two classes. After the intervention, we clearly can find that the black box replaced concept $2$ with a more representative visual concept of the stamen to \textit{water lily}.
	
In Figure \ref{fig:8}, RexNet100 misclassifies \textit{MD-87} as \textcolor{dgreen}{\textit{MD-80}}. The visually related intervention concept: \textcolor{dgreen}{\textit{"Large t-shaped  tail fin.", "white with blue and red stripes.", "large tail fin with the airline's logo.", etc.}} prompts us to delete the related concept about the tail with corresponding color-related and shape-related feature of aircraft to correct this error. We can find that the second most important visual concept extracted by the original black box before intervention corresponds exactly to this description. After the intervention, we can find that the network relies less on the tail of the aircraft to classify real class and the original score of the corresponding visual concept dropped from $0.45$ to $0.35$.
	
In Figure \ref{fig:9}, RexNet100 misclassifies \textit{white necked raven} as \textcolor{dgreen}{\textit{American crow}}. The visually related intervention concept: \textcolor{dgreen}{\textit{"Head is grayish-brown." and "large, black bird with a long, thick beak."}} prompts us to delete the related concept about the head and beak of bird to correct this error. We can find that the first most important visual concept extracted by the original black box before intervention corresponds exactly to this description. After the intervention, we can find that the network relies less on bird head to classify real class and the original score of the corresponding visual concept dropped from $0.37$ to $0.28$.

\end{document}